\documentclass[lettersize,journal]{IEEEtran}
\usepackage{amsmath,amsfonts}
\usepackage{algorithmic}
\usepackage{algorithm}
\usepackage{array}
\usepackage[caption=false,font=normalsize,labelfont=sf,textfont=sf]{subfig}
\usepackage{textcomp}
\usepackage{stfloats}
\usepackage{url}
\usepackage{verbatim}
\usepackage{graphicx}
\usepackage{cite}
\usepackage{times}  % DO NOT CHANGE THIS
\usepackage{helvet}  % DO NOT CHANGE THIS
\usepackage{courier}  % DO NOT CHANGE THIS
\usepackage{multirow}

\usepackage{amsmath}
\usepackage{amssymb}
\usepackage{xcolor}
\usepackage{amsthm}

\newtheorem{theorem}{Theorem}%[section]
\newtheorem{lemma}{Lemma}%[section]
\newtheorem{remark}{Remark}%[section]

\begin{document}

%\title{Suppressing Range of Adaptive Stepsizes in AdaBelief Improves Generalization of DNN Tasks}
\title{A DNN Optimizer that Improves over AdaBelief by Suppression of the Adaptive Stepsize Range}

\author{Guoqiang~Zhang,
        Kenta~Niwa,
        W.~Bastiaan~Kleijn
        % <-this % stops a space
\thanks{\IEEEcompsocthanksitem G.~Zhang is with the School of Electrical and Data Engineering, University of Technology, Sydney, Australia. Email:{guoqiang.zhang@uts.edu.au}}

\thanks{ K. Niwa is with both Communication Science Laboratories and Computer and Data Science Laboratories, Nippon Telegraph and Telephone Corporation (NTT). Email: {kenta.niwa.bk@hco.ntt.co.jp}} 

\thanks{ W.~Bastiaan~Kleijn is with Victory University of Wellington. Email:{bastiaan.kleijn@ecs.vuw.ac.nz }}

}



\maketitle

\begin{abstract}
We make contributions towards improving adaptive-optimizer performance. Our improvements are based on suppression of the range of adaptive stepsizes in the AdaBelief optimizer. Firstly,  we show that the particular placement of the parameter $\epsilon$ within the update expressions of AdaBelief reduces the range of the adaptive stepsizes, making AdaBelief closer to SGD with momentum.  Secondly, we extend AdaBelief by further suppressing the range of the adaptive stepsizes. To achieve the above goal, we perform mutual layerwise vector projections between the gradient $\boldsymbol{g}_t$ and its first momentum $\boldsymbol{m}_t$ before using them to estimate the second momentum.   The new optimization method is referred to as \emph{Aida}. Thirdly, extensive experimental results show that Aida outperforms nine optimizers when training transformers and LSTMs for NLP, and VGG and ResNet for image classification over CIAF10 and CIFAR100 while matching the best performance of the nine methods when training WGAN-GP models for image generation tasks. Furthermore, Aida produces higher validation accuracies than AdaBelief for training ResNet18 over ImageNet.  
\end{abstract}


%%%%%%%%%%%%%%%%%%%%%%%%%%%%%%%%%%%%%%%%%%%
\section{Introduction}
\label{sec:intro}

In the last decade, stochastic gradient descent (SGD) and its variants have been widely applied in deep learning \cite{Lecun15nature, Transformer17, Silver16GoGame, Chen20WaveGrad} due to their simplicity and effectiveness.  In the literature, SGD with momentum \cite{Sutskever13NAG, Polyak64CM}) dominates over other optimizers for image classification tasks \cite{He15ResNet, Wilson17AdamNegative}. Suppose the objective function $f({\boldsymbol{\theta}}):\boldsymbol{\theta}\in \mathbb{R}^d$ of a DNN model is differentiable. Its update expression for minimising $f(\boldsymbol{\theta})$ can be represented as
\begin{align}
\boldsymbol{m}_{t} & = \beta_t \boldsymbol{m}_{t-1} + \boldsymbol{g}_t  \label{equ:SGDM1}\\
\boldsymbol{\theta}_{t} &= \boldsymbol{\theta}_{t-1} - \eta_t \boldsymbol{m}_{t},  \label{equ:SGDM2}
\end{align}
where $\boldsymbol{g}_t = \nabla f(\boldsymbol{\theta}_{t-1})$ is the gradient at ${\boldsymbol{\theta}}_{t}$,  and $\eta_t$ is the common stepsize for all the coordinates of ${\boldsymbol{\theta}}$. In practice, the above method is often combined with a certain step-size scheduling method for $\eta_t$ when training DNNs.

To bring flexibility to SGD with momentum, an active research trend is to introduce elementwise adaptive stepsizes for all the coordinates of $\boldsymbol{m}_{t}$ in (\ref{equ:SGDM2}), referred to as \emph{adaptive optimization} \cite{Duchi11AdaGrad, Tieleman12RMSProp, Kingma17}. 
%\begin{figure}[t!]
%\centering
%\includegraphics[width=85mm]{Aida_K2_ALR_std_overall.eps}
%\includegraphics[width=140mm]{Aida_3stage_overall_std_mean.eps}
%\vspace*{-0.0cm}
%\caption{\footnotesize Layerwise standard deviations of elementwise stepsizes and layerwise average stepsizes for the top 10 neural layers %\footnotemark  
%when training VGG11 over CIFAR10 for 200 epochs. The jumps at 100 and 160 epoch in the curves are due to the change of the common stepsize. Adam+ is obtained by  replacing $(\boldsymbol{m}_t-\boldsymbol{g}_t)^2$ with $\boldsymbol{g}_t^2$ in AdaBelief where the parameter $\epsilon$ are placed differently from the original Adam (see the location differences of $\epsilon$ between (\ref{equ:Adam3}) and (\ref{equ:st_AdaBelief})-(\ref{equ:xt_AdaBelief}). The curves for Aida tend to be much more compact than those for AdaBelief and Adam+. See Fig.~\ref{fig:Aida_mean_std_compare} for more a concise comparison.}
%\label{fig:Aida_std_mean_compare}
%\vspace*{-0.3cm}
%\end{figure}
In the literature, Adam \cite{Kingma17} is probably the most popular adaptive optimization method (e.g., \cite{Transformer17,liu2021Swin,Zhang2019Adam,Devlin18Bert}). Its update expression can be written as 
\vspace{-0mm}
\begin{align}
\hspace{-10mm}[\textbf{Adam}]\hspace{0mm} &\left\{\begin{array}{l}
\hspace{-2mm}\boldsymbol{m}_{t}  =\beta_1\boldsymbol{m}_{t-1} + (1-\beta_1) \boldsymbol{g}_t      \qquad \qquad  \qquad  \quad \\
\hspace{-2mm}\boldsymbol{v}_{t}  = \beta_2 \boldsymbol{v}_{t-1} + (1-\beta_2) \boldsymbol{g}_t^2  \qquad  \qquad \qquad \quad   \\
%{x}_{t} &= {x}_{t-1} - \left\{\begin{array}{c} \eta \frac{\sqrt{1-\beta_2^t}}{1-\beta_1^t} \frac{{m}_{t}}{\sqrt{{v}_{t}+\epsilon}}   \\ 
% \eta \frac{\sqrt{1-\beta_2^t}}{1-\beta_1^t} \frac{{m}_{t}}{\sqrt{{v}_{t}}+\epsilon} 
%  \end{array}\right.  \label{equ:Adam3}
\hspace{-2mm} \boldsymbol{\theta}_{t}  = \hspace{-0.8mm} \boldsymbol{\theta}_{t-1}   \hspace{-0.8mm} - \hspace{-0.8mm}  \eta_t \frac{1}{1-\beta_1^{t}}  \frac{ \boldsymbol{m}_{t} }{\sqrt{\boldsymbol{v}_{t}/(1-\beta_2^{t})}+\epsilon}
\end{array}\right.\hspace{-20mm},
\label{equ:Adam}
\end{align}
where $\boldsymbol{g}_t=f(\boldsymbol{\theta}_{t-1})$, $0<\beta_1, \beta_2<1$, and $\epsilon>0$. The two vector operations $(\cdot)^2$ and $\cdot/\cdot$ are performed in an elementwise manner. The two exponential moving averages (EMAs) $\boldsymbol{m}_{t}$ and $\boldsymbol{v}_{t}$ are alternatively referred to as the first and second momentum. The two quantities $1-\beta_1^{t}$ and ${1-\beta_2^{t}}$ are introduced to compensate for the estimation bias in $\boldsymbol{m}_{t}$ and $\boldsymbol{v}_{t}$, respectively. $\eta_t$ is the common stepsize while $1/(\sqrt{\boldsymbol{v}_t/(1-\beta_2^{t})}+\epsilon)\in \mathbb{R}^d$ represents the elementwise adaptive stepsizes. %and the constant $\epsilon>0$ is introduced to avoid division by zero. $m_t$ and $v_t$ are the two moments of gradients and squared-gradients over iterations, respectively.  The two quantities $1-\beta_1^{t}$ and $\sqrt{1-\beta_2^{t}}$  in (\ref{equ:Adam3}) are introduced as bias-correction to compensate for the estimation bias in $m_{t}$ and $v_{t}$, respectively.  %Note that the update for $x_{t+1}$ may have different forms of expression depending on how the parameter $\epsilon$ participates in computation of the individual learning rates. As indicated by (\ref{equ:Adam3_1}) for example, the update direction $\frac{|{m}_{t+1}|\odot \textrm{sign}(m_{t+1}) }{\sqrt{{v}_{t+1}+\epsilon}}$ for $x_{t+1}$ can be viewed as the sign vector $\textrm{sign}(m_{t+1})$ multiplied element-wise by the magnitude vector $\frac{|{m}_{t+1}|}{\sqrt{{v}_{t+1}+\epsilon}}$.  

Due to the great success of Adam in training DNNs, various extensions of Adam have been proposed, including AdamW \cite{Loshchilov19AdamW}, NAdam \cite{Dozat16NAdam}, Yogi \cite{Zaheer18Yogi}, MSVAG  \cite{Balles17MSVAG}, Fromage \cite{Bernstein20Fromage}, and AdaBelief \cite{Zhuang20Adabelief}. It is worth noting that in \cite{Liu19RAdam}, the authors found that better generalization could be achieved by reducing the variance of the adaptive stepsizes of Adam. In doing so, they suggested multiplying a rectified scalar by $\boldsymbol{m}_t$ when computing $\boldsymbol{\theta}_t$ in (\ref{equ:Adam}) when the variance is large, which is referred to as RAdam. The AdaBound method of \cite{Luo19AdaBound} is designed to avoid extremely large and small adaptive stepsizes of Adam, which has a similar effect as RAdam. In practice, AdaBound works as an adaptive method at the beginning of the training process and gradually transforms to SGD with momentum, where all the adaptive stepsizes tend to converge to a single value. Conceptually speaking, both RAdam and AdaBound aim to reduce the range of the adaptive stepsizes of Adam to mimic the convergence behavior of SGD with momentum to a certain extent.  See also \cite{Wilson17AdamNegative} for a discussion on the relationship between the lack of generalization performance of adaptive methods and extreme adaptive stepsizes.  

Inspired by the above work \cite{Liu19RAdam,Luo19AdaBound}, we consider suppressing the range of adaptive stepsizes of AdaBelief. It is noted that AdaBelief extends Adam by tracking the EMA of the squared prediction error $(\boldsymbol{m}_t-\boldsymbol{g}_t)^2$. The update expressions of AdaBelief are given by 
\begin{align}
\hspace{-3mm}[\textbf{AdaBelief}]\hspace{0mm}&\left\{\begin{array}{l}
\hspace{-2mm}\boldsymbol{m}_t = \beta_1\boldsymbol{m}_{t-1}\hspace{-0.5mm} +\hspace{-0.5mm} (1\hspace{-0.5mm}-\hspace{-0.5mm}\beta_1)\boldsymbol{g}_t \\
\hspace{-2mm}\boldsymbol{s}_t=\beta_2 \boldsymbol{s}_{t-1}\hspace{-0.6mm}+\hspace{-0.6mm}(1\hspace{-0.6mm}-\hspace{-0.6mm}\beta_2)(\boldsymbol{m}_t\hspace{-0.6mm}-\hspace{-0.6mm}\boldsymbol{g}_t)^2+\textcolor{blue}{\epsilon}  \\
\hspace{-2mm}\boldsymbol{\theta}_t = \boldsymbol{\theta}_{t-1}  -\eta_t\frac{1}{1-\beta_1^t}\frac{\boldsymbol{m}_t}{\sqrt{\boldsymbol{s}_t/(1-\beta_2^t)}+\textcolor{blue}{\epsilon}}
\end{array}\right.\hspace{-2mm}.
\label{equ:AdaBelief}
\end{align}
We emphasise that the parameter $\epsilon$ is involved in the computation of both $\boldsymbol{s}_t$ and $\boldsymbol{\theta}_t$ in AdaBelief, which is different from that of Adam. %It remains unclear what the impact of $\epsilon$. 

%In general, each element in  $(\boldsymbol{m}_t-\boldsymbol{g}_t)^2$ would be smaller than the corresponding one in  $\boldsymbol{g}_t^2$ since $\boldsymbol{m}_t$ provides a reliable prediction of $\boldsymbol{g}_t$. As a result, the elementwise stepsizes of AdaBelief would be larger than those in Adam. 

In this paper, we make three contributions. Firstly, we explain why it is important to include the parameter $\epsilon$ in the computation of $\boldsymbol{s}_t$ in (\ref{equ:AdaBelief}), which will be inherited by our new algorithm Aida as described later on. In \cite{Zhuang20Adabelief}, the authors motivate the EMA of $(\boldsymbol{m}_t-\boldsymbol{g}_t)^2$  without explaining the inclusion of $\epsilon$. We show via a Taylor expansion that the inclusion of $\epsilon$ in the computation of $\boldsymbol{s}_t$ essentially suppresses the range of the adaptive stepsizes of AdaBelief. The above property makes AdaBelief closer to SGD with momentum.  %The above property provides one additional reason why AdaBelief performs better than Adam, and is consistent with the findings in \cite{Liu19RAdam,Luo19AdaBound}. 

\begin{figure}[t!]
\centering
\includegraphics[width=80mm]{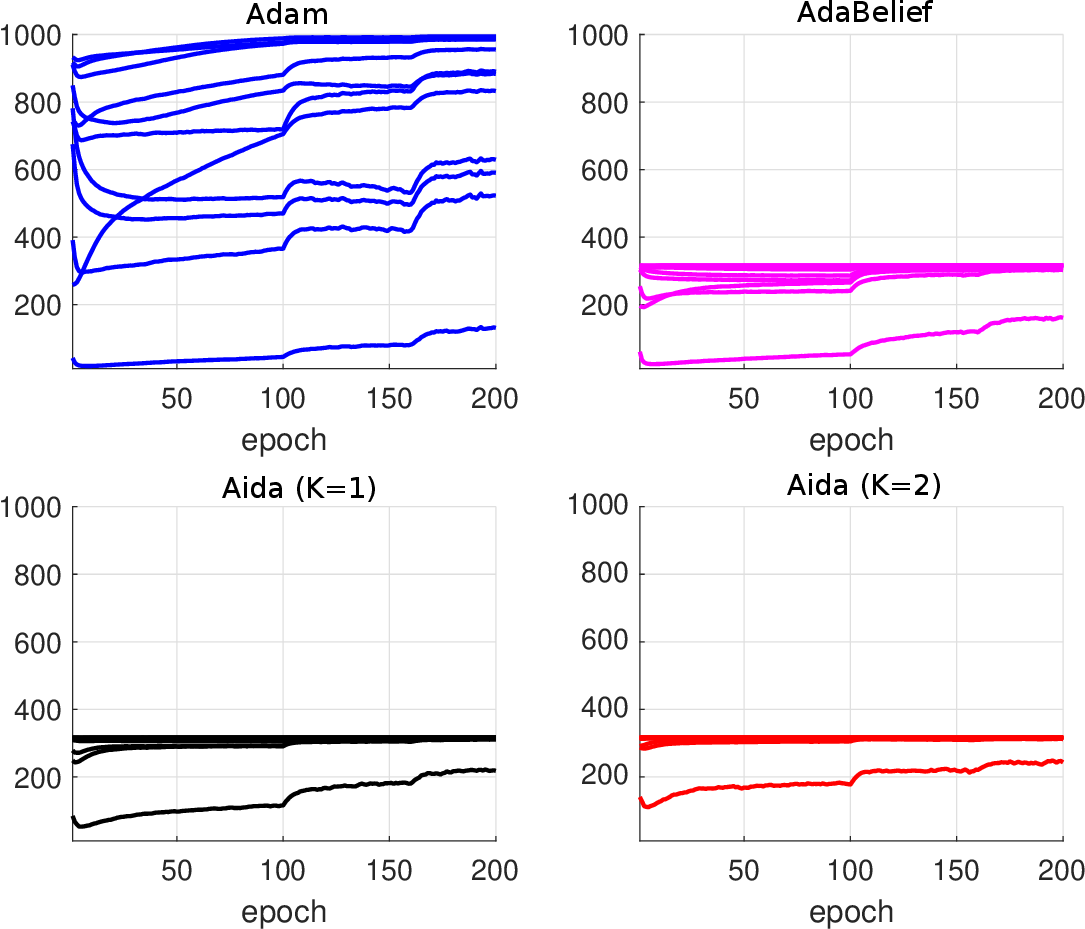}
\vspace*{-0.0cm}
\caption{\footnotesize Comparison of layerwise average of adaptive stepsizes for the 11 neural layers of VGG11 by training over CIFAR10 for 200 epochs. See Appendix~\ref{appendix:fig_setup} for the parameter setups of the three methods, where \textcolor{blue}{the optimal parameter $\epsilon$ for Adam was selected from a discrete set to give the best validation performance}. The jumps in the curves at 100 and 160 epochs are due to the change in the common stepsize. Aida has a much more compact range of layerwise average stepsizes than Adam and AdaBelief, respectively.   }
\label{fig:Aida_mean_compare}
\vspace*{-0.3cm}
\end{figure}

\footnotetext{As an example, considering Adam at iteration $t$, the layerwise average of adaptive stepsizes for the $l$th layer of VGG11 is computed as $\frac{1}{d_l}\sum_{i=1}^{d_l}1/(\sqrt{\boldsymbol{v}_{l,t}[i]/(1-\beta_2^t)}+\epsilon)$, where $\boldsymbol{v}_{l,t}\in \mathbb{R}^{d_l}$ is the subvector of the 2nd momentum $\boldsymbol{v}_{t}\in\mathbb{R}^d$ for the $l$th layer. }

\begin{figure}[t!]
\centering
\includegraphics[width=80mm]{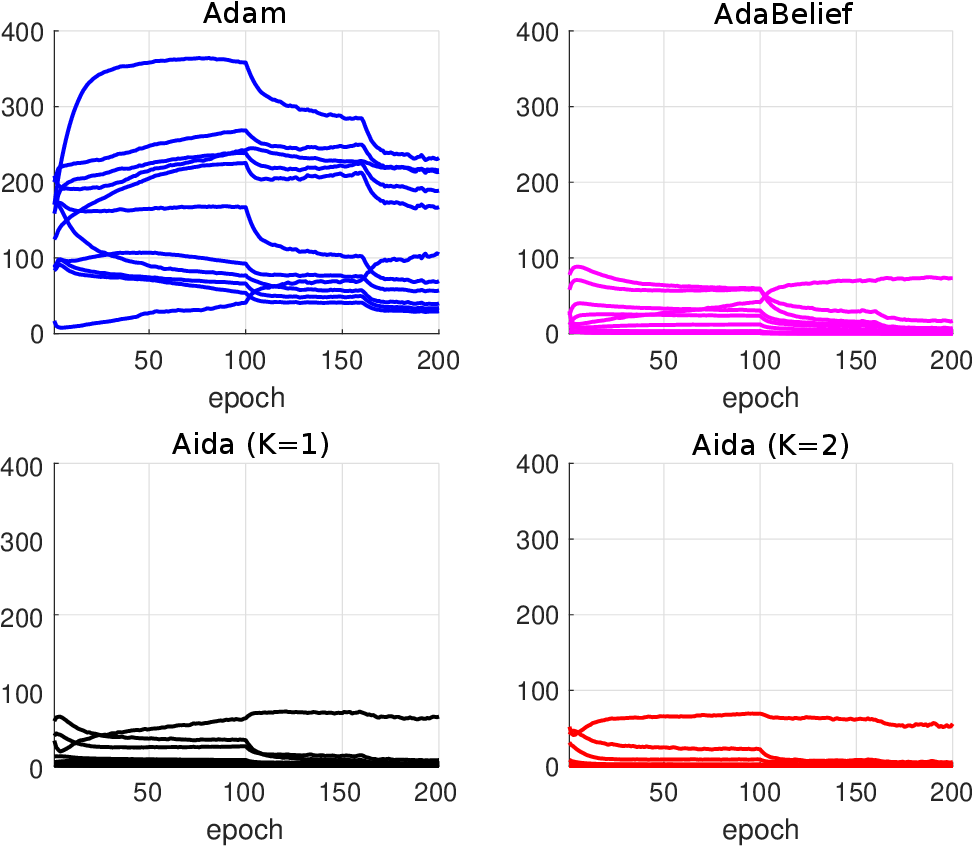}
\vspace*{-0.0cm}
\caption{\footnotesize{Comparison of layerwise standard deviations (stds) of adaptive stepsizes for the 11 neural layers by training VGG11 over CIFAR10 for 200 epochs. Aida has much smaller layerwise stds than Adam and AdaBelief, respectively.    } }
\label{fig:Aida_std_compare}
\vspace*{-0.3cm}
\end{figure}

Secondly, we perform layerwise vector projections to further suppress the range of adaptive stepsizes of AdaBelief. Let us denote the subvectors of $(\boldsymbol{m}_t, \boldsymbol{g}_t)$ for the $l$th layer of a DNN model as  $(\boldsymbol{m}_{l,t}, \boldsymbol{g}_{l,t})$. We perform $K$ mutual vector projections to obtain $(\boldsymbol{m}_{l,t}^{(K)}, \boldsymbol{g}_{l,t}^{(K)})$ for the $l$th layer starting from $(\boldsymbol{m}_{l,t}^{(0)}, \boldsymbol{g}_{l,t}^{(0)})=(\boldsymbol{m}_{l,t}, \boldsymbol{g}_{l,t})$. As an extension of AdaBelief, we then track and employ the EMA (or equivalently the second momentum) of $(\boldsymbol{m}_{l,t}^{(K)}-\boldsymbol{g}_{l,t}^{(K)})^2$ for the $l$th layer, where the resulting method is referred to \emph{Aida}.\footnote{It is named after an Italian opera by Verdi.} The new method has the nice property that the adaptive stepsizes within each neural layer have smaller statistical variance, and the layerwise average of the adaptive stepsizes are more compact across all the neural layers than the reference method. Detailed analysis will be provided later on.  As an example, Fig.~\ref{fig:Aida_mean_compare} and \ref{fig:Aida_std_compare} demonstrate that Aida indeed produces a more compact range of adaptive stepsizes than AdaBelief and Adam  for training VGG11 over CIFAR10. Furthermore, the adaptive stepsizes of Aida become increasingly more compact as the iteration increases. It is worth noting that at the end of the training process, the 11 layerwise average stepsizes in Fig.~\ref{fig:Aida_mean_compare}  do not converge to a single value, indicating the adaptability of Aida.  %This indicates that the behavior of Aida is closer to that of SGD with momentum than the other two methods. % in that the adaptive stepsizes of Aida have a more compact range.

Thirdly, extensive experimental results show that Aida with $K=2$ yields considerably better performance than nine optimization methods for training transformer \cite{Transformer17} and LSTM \cite{HochSchm97LSTM} models in natural language processing (NLP) tasks, and VGG11 \cite{Simonyan14VGG} and ResNet34 in image classification tasks over CIFAR10 and CIFAR100. It is also found that Aida matches the best performance of the nine methods when training WGAN-GP models in image generation tasks. Lastly, Aida outperforms AdaBelief when training ResNet18 on the large ImageNet dataset.  

The computational complexity of Aida was evaluated for training VGG11 and ResNet34. The results show that Aida with $K=2$ consumes an additional $25\%$ time per epoch compared to AdaBebelief. 

 %motivated by the observations in the first step, we propose to track the EMA of the \emph{layer-wise mean} value of square-gradient or its variants and then use them to compute the layer-wise adaptive step-sizes accordingly. We refer to the new layer-wise adaptive optimization framework as Aida.  Alg.~2 summarises the update procedure of Aida-grad which tracks the statistics of layer-wise squared-gradient over iterations (see Alg.~4 in Appendix~\ref{apsecA_belief} for Aida-belief as an extension of AdaBelief). 
%Convergence results are provided for Aida which follows almost the same analysis in \cite{Zhuang20Adabelief} for AdaBelief. 

\noindent\textbf{Notations}: We use small bold letters to denote vectors.  The $l_2$ and $l_{\infty}$ norms of a vector $\boldsymbol{y}$ are denoted as $\|\boldsymbol{y}\|_2$ and $\|\boldsymbol{y}\|_{\infty}$, respectively. Given an $L$-layer DNN model $\boldsymbol{\theta}$ of dimension $d$, we use $\boldsymbol{\theta}_l$ of dimension $d_l$ to denote the subvector of $\boldsymbol{\theta}$ for the $l$th layer. Thus, there is $\sum_{l=1}^L d_l=d$. The $i$th element of $\boldsymbol{\theta}_l$ is represented by $\boldsymbol{\theta}_l[i]$. The notation $[L]$ stands for the set $[L] = \{1,\ldots, L\}$. Finally, the angle between two vectors $\boldsymbol{y}$ and $\boldsymbol{x}$ of the same dimension is denoted by $\angle \boldsymbol{x}\boldsymbol{y}$. 

\section{Impact of $\epsilon$ in computation of $\boldsymbol{s}_t$ in AdaBelief}
By inspection of (\ref{equ:AdaBelief}), one can see that the parameter $\epsilon$ appears twice in the update expressions, the first one for computing $\boldsymbol{s}_t$ and the second one for computing $\boldsymbol{\theta}_t$. The impact of the second $\epsilon$ can be ignored due to the fact that $\sqrt{\epsilon/(1-\beta_2^t)} \gg \epsilon$ when $\epsilon$ is sufficiently small (e.g., $\epsilon=1e-8$). As a result, we only need to focus on the first $\epsilon$ when computing $\boldsymbol{s}_t$.

Next, we show that the first $\epsilon$ in the computation of $\boldsymbol{s}_t$ helps to suppress the range of adaptive stepsizes of AdaBelief. To this purpose, we reformulate the update expressions in (\ref{equ:AdaBelief}) as
\begin{align}
&\left\{\begin{array}{l}
\hspace{-2mm}\boldsymbol{m}_t = \beta_1\boldsymbol{m}_{t-1}\hspace{-0.5mm} +\hspace{-0.5mm} (1\hspace{-0.5mm}-\hspace{-0.5mm}\beta_1)\boldsymbol{g}_t \\
\hspace{-2mm}\hat{\boldsymbol{s}}_t=\beta_2 \hat{\boldsymbol{s}}_{t-1}\hspace{-0.6mm}+\hspace{-0.6mm}(1\hspace{-0.6mm}-\hspace{-0.6mm}\beta_2)(\boldsymbol{m}_t\hspace{-0.6mm}-\hspace{-0.6mm}\boldsymbol{g}_t)^2  \\
\hspace{-2mm}r_t = \beta_{2}r_{t-1}+\epsilon \\
\hspace{-2mm}\boldsymbol{\theta}_t = \boldsymbol{\theta}_{t-1}  -\eta_t\frac{1}{1-\beta_1^t}\frac{\boldsymbol{m}_t}{\sqrt{(\hat{\boldsymbol{s}}_t+r_t)/(1-\beta_2^t)}}
\end{array}\right.\hspace{-2mm},
\label{equ:AdaBelief2}
\end{align}
where the second $\epsilon$ is removed, and 
\begin{align}
\boldsymbol{s}_t=\hat{\boldsymbol{s}}_t+r_t = \hat{\boldsymbol{s}}_t+\epsilon(1-\beta_2^t)/(1-\beta_2), \label{equ:r_t} \end{align}
where $\hat{\boldsymbol{s}}_0=0$ and $r_0 = 0$. 
As a result, the adaptive stepsizes $1/\sqrt{(\hat{\boldsymbol{s}}_t+r_t)/(1-\beta_2^t)}$ in (\ref{equ:AdaBelief2}) can be approximated to be
\begin{align}
&1/\sqrt{(\hat{\boldsymbol{s}}_t+r_t)/(1-\beta_2^t)} \nonumber \\
&= 1/\sqrt{\hat{\boldsymbol{s}}_t/(1-\beta_2^t)+\epsilon/(1-\beta_2)} \label{equ:epsilon_effect0} \\
&\approx  \frac{1}{\underbrace{\sqrt{\hat{\boldsymbol{s}}_t/(1-\beta_2^t)}}_{\textrm{1st term}}+\underbrace{\frac{1}{2\sqrt{\hat{\boldsymbol{s}}_t/(1-\beta_2^t)}}\epsilon/(1-\beta_2)}_{\textrm{2nd term}} },\label{equ:epsilon_effect}
\end{align}
where in the last step, the Taylor approximation is applied to a function $h(x)=\sqrt{\boldsymbol{a}+x}$ around $x=0$, where $x=\epsilon/(1-\beta_2)$ and $\boldsymbol{a}=\hat{\boldsymbol{s}}_t/(1-\beta_2^t)$.

\begin{figure}[t!]
\centering
\includegraphics[width=80mm]{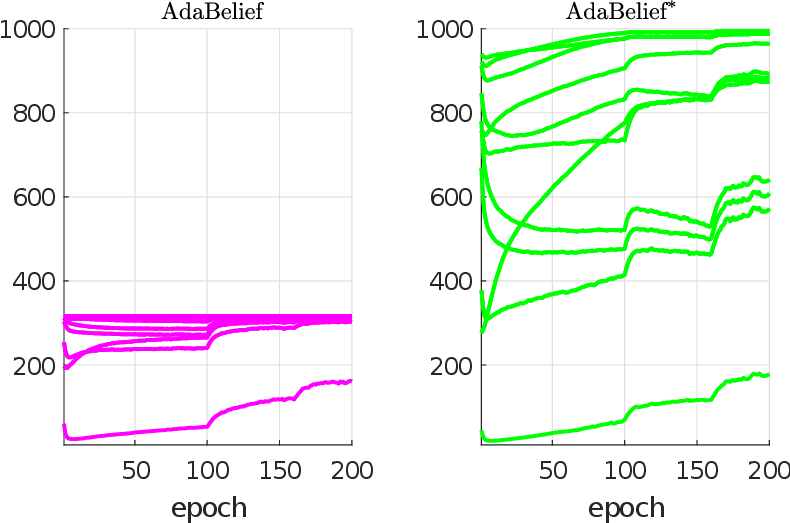}
\vspace*{-0.2cm}
\caption{\footnotesize{Comparison of layerwise average of adaptive stepsizes for the 11 neural layers by training VGG11 over CIFAR10 for 200 epochs. AdaBelief$^*$ is obtained by removing the first $\epsilon$ in  the computation of $\boldsymbol{s}_t$ and keeping the second $\epsilon$. See Appendix~\ref{appendix:AdaBeliefstar} and \ref{appendix:fig_setup} for the update procedure of AdaBelief* and the parameter setups of the two optimization methods. \textcolor{blue}{The optimal $\epsilon$ for AdaBelief* is selected from a discrete set that gives the best validation accuracy. } } }
\label{fig:AdaBeliefMinus}
\vspace*{-0.3cm}
\end{figure}

%We now investigate (\ref{equ:epsilon_effect}). Generally speaking, large elements of $\hat{\boldsymbol{s}}_t$ would lead to small adaptive stepsizes while small elements lead to large adaptive stepsizes due to the inverse operation $1/(\cdot)$. It is clear from (\ref{equ:epsilon_effect}) that for large elements of $\hat{\boldsymbol{s}}_t$, the second term in the denominator of (\ref{equ:epsilon_effect}) is relatively small, and therefore have little effect on the resulting small adaptive stepsizes. In contrast, for small elements of $\hat{\boldsymbol{s}}_t$, the second term is relatively large, thus avoiding extremely large adaptive stepsizes. To briefly summarize, the inclusion of $\epsilon$ in computation of $\boldsymbol{s}_t$ suppresses the range of adaptive stepsizes in AdaBelief  by avoiding extremely large stepsizes. 

We now investigate (\ref{equ:epsilon_effect}). Generally speaking,  small elements of $\hat{\boldsymbol{s}}_t$ lead to large adaptive stepsizes while large elements lead to small adaptive stepsizes due to the inverse operation $1/(\cdot)$. It is clear from (\ref{equ:epsilon_effect}) that for small elements of $\hat{\boldsymbol{s}}_t$, the second term in the denominator is relatively large, implicitly penalizing large stepsizes. Furthermore, (\ref{equ:epsilon_effect0}) indicates that those large stepsizes are upper-bounded by the quantity $1/\sqrt{\epsilon/(1-\beta_2)}$.  In contrast, for large elements of $\hat{\boldsymbol{s}}_t$, the second term is relatively small, thus avoiding extremely small adaptive stepsizes. In short, including $\epsilon$ in the computation of $\boldsymbol{s}_t$ suppresses the range of adaptive stepsizes in AdaBelief by avoiding extremely small stepsizes.

Fig.~\ref{fig:AdaBeliefMinus} demonstrates that when the first $\epsilon$ is removed from (\ref{equ:AdaBelief}) in AdaBelief, the resulting method AdaBelief$^*$ indeed has a broader range of adaptive stepsizes than AdaBelief.  %Furthermore, the largest layerwise average stepsize of AdaBelief is around 300 while that of AdaBelief$^*$ is around 1000. 
At epoch 200, the eleven layerwise average stepsizes in AdaBelief$^*$ are distributed in [190,1000] while ten out of eleven layerwise average stepsizes in AdaBelief are close to a single value of 320. 
That is, the first $\epsilon$ in (\ref{equ:AdaBelief}) indeed makes the adaptive sizes of AdaBelief more compact. %The respective validation accuracy for each method is also included in Fig. ~\ref{fig:AdaBeliefMinus}, which is obtained by three independent experimental repetitions.  It shows that smaller range of adaptive stepsizes in AdaBelief leads to higher validation accuracy, which is consistent with the findings of \cite{Liu19RAdam,Luo19AdaBound}.  %As are presented in Algorithm 1, our two new methods \emph{Aida} and \emph{AidaGrad} also include $\epsilon$ in computa $\emph{AidaGrad}$ .      

\begin{remark}
To reduce the range of adaptive stepsizes in Adam, one can also include $\epsilon$ in computing $\boldsymbol{v}_t$ in (\ref{equ:Adam}). See Fig.~\ref{fig:AdamPlus} in Appendix~\ref{appendix:Adamplus_exp} for demonstration when training VGG11 over CIFAR10.   
\end{remark}

\vspace{-2mm}
\section{Algorithmic Design }
\label{sec:alg}
\vspace{-1mm}
We showed in the  previous section that the particular placement of $\epsilon$ in the update expressions of AdaBelief suppresses the range of the adaptive stepsizes. In this section, we develop a new technique to further reduce the range of adaptive stepsizes of AdaBelief, which is referred to as \emph{layerwise vector projections}. The new method is named \emph{Aida}. A convex convergence analysis is presented at the end of the section.    

%In this section, we first  motivate a general strategy that could suppress the range of adaptive stepsizes of Adam and AdaBelief. After that, we present the update expressions of Aida and AidaGrad.  Convex convergence analysis is presented in the end.       

%In this subsection, we first explain in detail why our new method Aida is ought to track a quantity of $\boldsymbol{m}_{l,t}$ and $\boldsymbol{g}_{l,t}$ for the $l$th layer of a DNN model with smaller statistic variance and average value over all the coordinates than $(\boldsymbol{m}_{l,t}-\boldsymbol{g}_{l,t})^2$ being employed in AdaBelief. After that, we present the update expressions of Aida.  Convex convergence analysis is presented in the end. 

\vspace{-2mm}
\subsection{Motivation}
%\noindent\textbf{Motivation}:
Our aim is to design a new adaptive optimization algorithm, in which the range of the adaptive stepsizes is smaller than that of AdaBelief. To achieve the above goal, we consider processing $\boldsymbol{m}_t$ and $\boldsymbol{g_t}$ in a layerwise manner at iteration $t$ before using them to estimate the second momentum $\boldsymbol{v}_t$. Due to the nature of back-propagation when training a DNN model, it is computationally more efficient to perform layerwise processing than operating on the entire vector $\boldsymbol{m}_t$ and $\boldsymbol{g}_t$. On the other hand, the parameters within the same neural layer are functionally homogeneous when processing data from the layer below. It is likely that the gradients within the same layer follow a single distribution, making it natural to perform layerwise processing.

We now consider what kind of layerwise processing would be desirable to extend AdaBelief. Firstly, we note that the parameter $\epsilon$ of (\ref{equ:AdaBelief}) in AdaBelief essentially defines an upper bound on the adaptive stepsizes and is independent of neural layer and iteration indices. By inspection of (\ref{equ:AdaBelief2})-(\ref{equ:epsilon_effect0}), the upper bound can be expressed as
\begin{align}
\frac{1}{\sqrt{r_t/(1-\beta_2^t)}+\epsilon} = \frac{1}{\sqrt{\epsilon/(1-\beta_2)}+\epsilon}.
\label{equ:up_bound_AdaBelief}
\end{align}
We use $(\boldsymbol{m}_{l,t}, \boldsymbol{g}_{l,t})$ to denote the subvectors of $(\boldsymbol{m}_{t}, \boldsymbol{g}_{t})$ for the $l$th neural layer. Suppose we track the EMA of $(\gamma_{l,t}\boldsymbol{m}_{l,t} - \beta_{l,t}\boldsymbol{g}_{l,t})^2$ for the $l$th layer, where $0<\gamma_{l,t}, \beta_{l,t}\leq 1$ are functions of $(\boldsymbol{m}_{l,t}, \boldsymbol{g}_{l,t})$, instead of the EMA of $(\boldsymbol{m}_{l,t}-\boldsymbol{g}_{l,t})^2$ being tracked in AdaBelief. If the scalars $\{\gamma_{l,t},\beta_{l,t}\}$ are sufficiently small in the extreme case, all the adaptive stepsizes of the new method tend to approach the upper bound in (\ref{equ:up_bound_AdaBelief}). As a result, the new method will have a smaller range of adaptive stepsizes than AdaBelief either in a layerwise manner or globally. %Similarly, Adam can be extended by tracking the EMA of $(\beta_{l,t}\boldsymbol{g}_{l,t})^2$ for the $l$th layer.   

We propose to compute the scalars $(\gamma_{l,t},\beta_{l,t})$ of $\boldsymbol{m}_{l,t}$ and $\boldsymbol{g}_{l,t}$ mentioned above via $K$ mutual vector projections starting from $(\boldsymbol{m}_{l,t},\boldsymbol{g}_{l,t})$ (see Fig.~\ref{fig:vector_projection} for demonstration). In practice, it is found that $K=2$ is sufficient to produce small scalars $(\gamma_{l,t},\beta_{l,t})$, leading to a smaller range of adaptive stepsizes than those of AdaBelief and Adam. The parameter $K$ of Aida in Fig.~\ref{fig:Aida_mean_compare}-\ref{fig:Aida_std_compare} was set to $K\in \{1,2\}$. In the following, the update expressions of Aida are presented in detail.

\subsection{Aida as an extension of AdaBelief}
\begin{figure}[t!]
\centering\includegraphics[width=80mm]{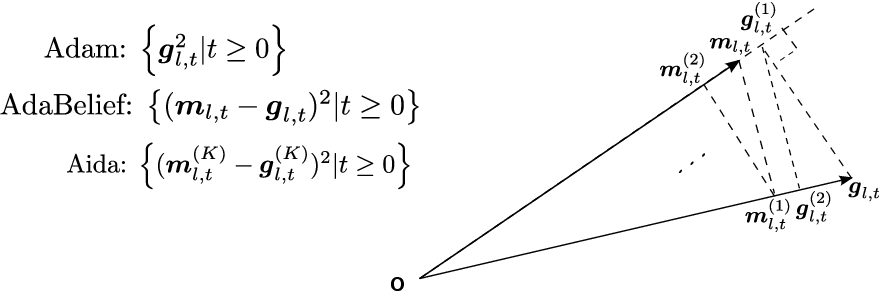}
\vspace*{-0.0cm}
\caption{\footnotesize{ Computation of $\{(\boldsymbol{m}_{l,t}^{(k)},\boldsymbol{g}_{l,t}^{(k)})|k= [K] \}$ by starting from the pair $(\boldsymbol{m}_{l,t},\boldsymbol{g}_{l,t})$ via sequential and alternating vector projections in Aida.} }
\label{fig:vector_projection}
\vspace*{-0.3cm}
\end{figure}

Consider the $l$th layer of a DNN model at iteration $t$. %We would like to track the EMA of a more conservative function of $\boldsymbol{m}_{l,t}$ and $\boldsymbol{g}_{l,t}$ than the squared prediction error $(\boldsymbol{m}_{l,t}-\boldsymbol{g}_{l,t})^2$ as employed in AdaBelief. %We note that in principle, various functions may be constructed to achieve the above goal.
We perform a sequence of mutual vector projections to obtain a set of projected vectors $\{ (\boldsymbol{m}_{l,t}^{(k)}, \boldsymbol{g}_{l,t}^{(k)}) |\, k=[K]\}$ starting from the initial pair  $(\boldsymbol{m}_{l,t}^{(0)}, \boldsymbol{g}_{l,t}^{(0)}) =(\boldsymbol{m}_{l,t}, \boldsymbol{g}_{l,t})$. Using algebra, the two vectors at iteration $k$ can be represented as \begin{align}
\boldsymbol{m}_{l,t}^{(k+1)} &=  \frac{\langle \boldsymbol{g}_{l,t}^{(k)}, \boldsymbol{m}_{l,t}^{(k)}\rangle }{\|\boldsymbol{g}_{l,t}^{k}\|_2^2+\xi} \boldsymbol{g}_{l,t}^{(k)} \label{equ:g_m_proj} \\
\boldsymbol{g}_{l,t}^{(k+1)} &=  \frac{\langle \boldsymbol{g}_{l,t}^{(k)}, \boldsymbol{m}_{l,t}^{(k)}\rangle}{\|\boldsymbol{m}_{l,t}^{(k)}\|_2^2+\xi} \boldsymbol{m}_{l,t}^{(k)}, \label{equ:m_g_proj}
\end{align}
where $\langle\cdot, \cdot \rangle$ denotes the inner product, and $\xi>0$ is a scalar parameter to make sure the division operations are valid. The above two projections (\ref{equ:g_m_proj})-(\ref{equ:m_g_proj}) ensure that the resulting projected vectors share the same vector-direction as either $\boldsymbol{m}_{l,t}$ or $\boldsymbol{g}_{l,t}$. See Fig.~\ref{fig:vector_projection} for visualisation.

Once $(\boldsymbol{m}_{l,t}^{(K)},\boldsymbol{g}_{l,t}^{(K)})$ are obtained for the $l$th layer, Aida tracks the EMA of the squared difference $(\boldsymbol{m}_{l,t}^{(K)} - \boldsymbol{g}_{l,t}^{K})^2$, given by 
\begin{align}
    \boldsymbol{v}_{l,t} = \beta_2 \boldsymbol{v}_{l,t-1} + (1-\beta_2) \left(\boldsymbol{m}_{l,t}^{(K)} - \boldsymbol{g}_{l,t}^{(K)}\right)^{2} + \epsilon, \label{equ:v_Aida}
\end{align}
where $1>\beta_2>0$, and $\epsilon>0$ is added as recommended by our earlier analysis. With $\boldsymbol{v}_{l,t}$, the model parameters $\boldsymbol{\theta}_{l,t}$ of the $l$th layer can be updated accordingly. See Algorithm~1 for a summary of Aida.

Next, we consider the geometric properties of the set of projected vectors.  It is not difficult to show that after projection, the resulting vectors have either shorter or equal length in comparison to the original vectors: \begin{align}
\|\boldsymbol{m}_{l,t}^{(k)}\|_2 \leq \|\boldsymbol{m}_{l,t}^{k-1}\|_2
\quad \textrm{and} \quad \|\boldsymbol{g}_{l,t}^{(k)}\|_2 \leq \|\boldsymbol{g}_{l,t}^{(k-1)}\|_2. \label{equ:proj_ineq} \end{align}
Using the fact that mutual projections of two vectors do not change the angle, we then have
\begin{align}
    \| \boldsymbol{m}_{l,t}^{(k)} - \boldsymbol{g}_{l,t}^{(k)} \|_2 \leq \| \boldsymbol{m}_{l,t}^{(k-1)} - \boldsymbol{g}_{l,t}^{(k-1)} \|_2, 
    \label{equ:proj_norm_ineq}
\end{align}
where the equality holds if $\boldsymbol{m}_{l,t}$ and $\boldsymbol{g}_{l,t}$ are on the same line and $\xi$ can be ignored in  (\ref{equ:m_g_proj}). 

For the extreme case that each neural layer has only one parameter (i.e., $\boldsymbol{g}_{l,t}\in \mathbb{R}$, $\forall l\in [L]$), it is easy to show that the projection operation has no effect. That is, $(\boldsymbol{g}_{l,t}^{(k)} - \boldsymbol{m}_{l,t}^{(k)})^2 =(\boldsymbol{g}_{l,t}-\boldsymbol{m}_{l,t})^2$ for all $k\in [K]$ if $\xi$ is ignored in  (\ref{equ:m_g_proj}). In this case, Aida reduces to AdaBelief.

From the above analysis, we can conclude that the EMA of $(\boldsymbol{m}_{l,t}^{(K)} - \boldsymbol{g}_{l,t}^{K})^2$ for the $l$th layer can be viewed as the EMA of $(\gamma_{K, l,t}\boldsymbol{m}_{l,t} - \beta_{K, l,t}\boldsymbol{g}_{l,t})^2$, where the scalars $\gamma_{K, l,t}, \beta_{K, l,t}\in (0,1]$. In general, the angles $\{\angle\boldsymbol{g}_{l,t} \boldsymbol{m}_{l,t} | t\geq 0 \}$ would be non-zero due to randomness introduced by the minibatch training strategy in a typical DNN task. As a result, increasing the number $K$ of vector projections would cause the elements of $\{\gamma_{K, l,t}, \beta_{K, l,t}| t\geq 0\}$ to approach zero. In other words, the parameter $K$ controls the range of the adaptive stepsizes of Aida. A larger $K$ makes the adaptive stepsizes more compact.

Figs.~\ref{fig:Aida_mean_compare} and \ref{fig:Aida_std_compare} provide empirical evidence that Aida does indeed have a smaller range of adaptive stepsizes than AdaBelief and Adam. Furthermore, as $K$ increases from 1 to 2, the range of adaptive stepsizes of Aida becomes increasingly compact. Hence, Aida is closer to SGD with momentum than AdaBelief. As will be demonstrated in the experiments, Aida improves the generalization of Adam and AdaBelief for several classical DNN tasks.

%Consider the extreme case that the histogram of all the adaptive stepsizes is quite broad and flat over iterations. This would make the impact of the gradient descent operations strongly nonuniform in the parameter space. Certain neurons with very small adaptive stepsizes may receive updates of relatively low magnitude during the entire training procedure, which likely is undesirable for the learning process.

\begin{algorithm}[t!]
   \caption{\small Aida: Suppressing the range of adaptive stepsizes of AdaBelief by layerwise vector projections}
   \label{alg:Adam}
\begin{algorithmic}[1]
   \STATE {\small {\bfseries Input:} $\beta_1$, $\beta_2$,  $\eta_t$, $\epsilon > 0$, $\xi=1e-20$, $K=2$  }
   \STATE {\small {\bfseries  Init.:} $\boldsymbol{\theta}_0\hspace{-0.5mm}\in\hspace{-0.5mm} \mathbb{R}^d$,  $\boldsymbol{m}_0 \hspace{-0.5mm}=\hspace{-0.5mm} 0$, $\boldsymbol{v}_{0}=\tilde{\boldsymbol{v}}_{0}=0  \in \mathbb{R}^d$ }
   \FOR{\small $t=1, 2, \ldots, T$}
   \STATE \hspace{-0mm}{\small  $\boldsymbol{g}_t \leftarrow \nabla f({\boldsymbol{\theta}}_{t-1}) $  }
   \STATE \hspace{-0mm}{\small $\boldsymbol{m}_{t} \leftarrow \beta_1 \boldsymbol{m}_{t-1}  + (1-\beta_1) \boldsymbol{g}_t$ }
   \FOR{\small $l=1,\ldots, L$}
   \STATE {\small $\boldsymbol{m}^{(0)}_t = \boldsymbol{m}^{(0)}_t,\;
   \boldsymbol{g}^{(0)}_t = \boldsymbol{g}_t$}
   \FOR{\small $k=1,\ldots,K$} 
   \STATE $\boldsymbol{m}_{l,t}^{(k)} = \frac{\langle \boldsymbol{m}_{l,t}^{k-1}, \boldsymbol{g}_{l,t}^{k-1} \rangle}{\|\boldsymbol{g}_{l,t}^{(k-1)}\|_2^2+\xi}\boldsymbol{g}_{l,t}^{(k-1)}$ 
 \STATE $ \boldsymbol{g}_{l,t}^{(k)} = \frac{\langle \boldsymbol{m}_{l,t}^{k-1}, \boldsymbol{g}_{l,t}^{k-1} \rangle}{\|\boldsymbol{m}_{l,t}^{(k-1)}\|_2^2+\xi}\boldsymbol{m}_{l,t}^{(k-1)} $
   \ENDFOR
    \STATE $\boldsymbol{v}_{l,t}\hspace{-0.6mm}\leftarrow\hspace{-2mm}\begin{array}{l}\hspace{-0mm} \beta_2\boldsymbol{v}_{l,t-1}\hspace{-0.6mm}+\hspace{-0.6mm}(1\hspace{-0.5mm}-\hspace{-0.5mm}\beta_2)(\boldsymbol{m}_{l,t}^{(K)}\hspace{-0.6mm}-\hspace{-0.6mm}\boldsymbol{g}_{l,t}^{(K)})^2\hspace{-0.6mm}+\hspace{-0.6mm}\epsilon %\;\;\;[\textbf{Aida}] %\\
   %\hspace{-2.5mm}\beta_2\boldsymbol{v}_{l,t-1}\hspace{-0.6mm}+\hspace{-0.6mm}(1\hspace{-0.5mm}-\hspace{-0.5mm}\beta_2)(\boldsymbol{g}_{l,t}^{(K)})^2\hspace{-0.6mm}+\hspace{-0.6mm}\epsilon \;\;\;\;\;\;\;[\textbf{AidaGrad}]
   \end{array}$
   \ENDFOR
   \STATE \hspace{-0mm}{\small $  \tilde{\boldsymbol{m}}_{t} \hspace{-0.6mm}\leftarrow \frac{\boldsymbol{m}_{t}}{1-\beta_1^{t}}\quad   \left\{\tilde{\boldsymbol{v}}_{l,t} \leftarrow \frac{\boldsymbol{v}_{l,t}}{1-\beta_2^{t}} \right\}_{l=1}^L $ } 
  \STATE {\small  $\boldsymbol{\theta}_{t} \hspace{-0.6mm}\leftarrow \boldsymbol{\theta}_{t-1} -\frac{\eta_t\tilde{\boldsymbol{m}}_{t} }{\sqrt{\tilde{\boldsymbol{v}}_{t}} } $}
   \ENDFOR 
   \STATE {\bfseries Output:} {\small $\boldsymbol{\theta}_{T}$  }
\end{algorithmic}
\end{algorithm} 
 
\subsection{Convergence analysis}

In this paper, we focus on convex optimization for Aida. Our analysis follows a strategy similar to that used to analyse AdaBelief in \cite{Zhuang20Adabelief}. Note that the upper bound we obtain is essentially tighter than that in \cite{Zhuang20Adabelief} due to two minor corrections.\footnote{1. As will be shown later, we do not replace $\beta_{1t}$ by $\beta_1$ when dealing with the quantity $\frac{1}{2\eta_t(1-\beta_{1t})}[\|\boldsymbol{v}_t^{1/4}(\boldsymbol{\theta}_{t-1}-\boldsymbol{\theta}^{\ast})\|_2^2 -\|\boldsymbol{v}_t^{1/4}(\boldsymbol{\theta}_{t}-\boldsymbol{\theta}^{\ast})\|_2^2 ]$ as is done in the derivation of (3) in the appendix of \cite{Zhuang20Adabelief}. We have added the dimensionality $d$ to the last quantity of (6) in the appendix of \cite{Zhuang20Adabelief}. } In particular, the first term in (\ref{equ:Aida_convex}) is of order $O(1/T)$ while the corresponding one in \cite{Zhuang20Adabelief} is essentially of order $1/(\sqrt{T})$.  In other words, we have improved the regret bound of \cite{Zhuang20Adabelief}.

\begin{theorem}
Suppose $\{\boldsymbol{\theta}_t\}_{t=0}^T$ and $\{\boldsymbol{v}_t\}_{t=0}^T$ are the iterative updates obtained by either Aida\footnote{$\beta_1$ in Algorithm~1 is generalized to be $\beta_{1t}$, $t\geq 0$ to facilitate convergence analysis.} starting with $(\boldsymbol{m}_0,\boldsymbol{v}_0)=(\boldsymbol{0},\boldsymbol{0})$. Let  $0\leq\beta_{1t}=\beta_1\lambda^t<1, 0\leq\beta_2<1$, and  $\eta_t =\frac{\eta}{\sqrt{t}}$. Assume (1): $f(\boldsymbol{\theta})$ is a differentiable convex function with $\|\boldsymbol{g}_t\|_{\infty}\leq G_{\infty}/2$ (hence $\|\boldsymbol{m}_{l,t}^{(K)}-\boldsymbol{g}_{l,t}^{(K)}\|_{\infty}\leq G_{\infty}$) for all $t\in [T]$; (2): the updates $\{\boldsymbol{\theta}_t\}_{t=0}^T$ and the optimal solution $\boldsymbol{\theta}^{\ast}$ are bounded by a hyper-sphere, i.e., $\|\boldsymbol{\theta}_t\|_2\leq D$ and $\|\boldsymbol{\theta}^{\ast}\|_2\leq D$; (3): $0<c\leq \boldsymbol{v}_{t-1}[i] \leq \boldsymbol{v}_{t}[i]$ for all $i\in \{1,\ldots, d\} $ and $t\in [T]$. Denote $\bar{\boldsymbol{\theta}}_{T}=\frac{1}{T}\sum_{t=0}^{T-1} \boldsymbol{\theta}_t$ and ${\boldsymbol{g}_{1:T}^2[i]}=((\boldsymbol{g}_1[i])^2,\ldots, (\boldsymbol{g}_T[i])^2)\in \mathbb{R}^{T}$. We then have the following bound on regret: 
{\small\begin{align}
&f(\bar{\boldsymbol{\theta}}_{T})-f(\boldsymbol{\theta}^{\ast})]\leq  \overbrace{\frac{D^2}{\eta(1-\beta_1)T}\sum_{i=1}^d(\boldsymbol{v}_{1}[i])^{1/2}}^{\textcolor{blue}{\textrm{1st term}}} \nonumber \\ &+\frac{(1+\beta_1)\eta\sqrt{1+\log{T}}}{2\sqrt{c}(1-\beta_1)^3 T}\sum_{i=1}^d\left\|\boldsymbol{g}_{1:T}^2[i]\right\|_2+ \frac{D^2\beta_1G_{\infty}d}{(1-\beta_1)(1-\lambda)^2\eta T}. \label{equ:Aida_convex}
\end{align}}
\label{theorem:convex}
\end{theorem}
\begin{proof}

%We now provide the derivations for the upper bound in the theorem. 

Firstly, we note that the two bias terms $1-\beta_1^t$ and $1-\beta_2^t$ in the update expressions of Aida in Algorithm~1 can be absorbed into the common stepsize $\eta_t$. Therefore, we will ignore the two bias terms in the following proof. 
%Secondly, the lower bound $0<c\leq \boldsymbol{v}_t[i]$%
Suppose $\boldsymbol{\theta}^{\ast}$ is the optimal solution for solving the convex optimization problem, i.e., $\theta^{\ast}=\arg\min_{\boldsymbol{\theta}_t} f(\boldsymbol{\theta})$. Using the fact that $\boldsymbol{\theta}_t = \boldsymbol{\theta}_{t-1}-\eta_t \boldsymbol{v}_t^{-1/2}\boldsymbol{m}_t$, we have
\begin{align}
&\|\boldsymbol{v}_t^{1/4}(\boldsymbol{\theta}_t-\boldsymbol{\theta}^{\ast}) \|_2^2 \nonumber \\
&= 
\|\boldsymbol{v}_t^{1/4}(\boldsymbol{\theta}_{t-1}-\eta_t\boldsymbol{v}_{t}^{-1/2}\boldsymbol{m}_t-\boldsymbol{\theta}^{\ast})  \|_2^2 \nonumber \\
&=\|\boldsymbol{v}_t^{1/4}(\boldsymbol{\theta}_{t-1}-\boldsymbol{\theta}^{\ast}) \|_2^2+\eta_t^2\|\boldsymbol{v}_t^{-1/4}\boldsymbol{m}_t\|_2^2 \nonumber \\
& \hspace{5mm}- 2\eta_t\langle \beta_{1t}\boldsymbol{m}_{t-1}+(1-\beta_{1t})\boldsymbol{g}_t, \boldsymbol{\theta}_{t-1}-\boldsymbol{\theta}^{\ast} \rangle \nonumber \\
&=\|\boldsymbol{v}_t^{1/4}(\boldsymbol{\theta}_{t-1}-\boldsymbol{\theta}^{\ast}) \|_2^2+\eta_t^2\|\boldsymbol{v}_t^{-1/4}\boldsymbol{m}_t\|_2^2 \hspace{-0.6mm} \nonumber \\
&\hspace{5mm}-\hspace{-0.6mm} 2\eta_t(1\hspace{-0.6mm}-\hspace{-0.6mm}\beta_{1t})\langle \boldsymbol{g}_t, \boldsymbol{\theta}_{t-1}\hspace{-0.6mm}-\hspace{-0.6mm}\boldsymbol{\theta}^{\ast} \rangle \hspace{-0.6mm}-\hspace{-0.6mm} 2\eta_t\beta_{1t}\langle \boldsymbol{m}_{t-1},  \boldsymbol{\theta}_{t-1}\hspace{-0.6mm}-\hspace{-0.6mm}\boldsymbol{\theta}^{\ast} \rangle \nonumber  \\
&\hspace{-0mm}\leq \|\boldsymbol{v}_t^{1/4}(\boldsymbol{\theta}_{t-1}-\boldsymbol{\theta}^{\ast}) \|_2^2+\eta_t^2\|\boldsymbol{v}_t^{-1/4}\boldsymbol{m}_t\|_2^2 \hspace{-0.6mm} \nonumber \\
& \hspace{5mm}-\hspace{-0.6mm} 2\eta_t(1-\beta_{1t})\langle \boldsymbol{g}_t, \boldsymbol{\theta}_{t-1}-\boldsymbol{\theta}^{\ast} \rangle \hspace{-0.6mm} \nonumber \\
& \hspace{5mm}+ \eta_t^2\beta_{1t}\|\boldsymbol{v}_t^{-1/4}\boldsymbol{m}_{t-1}\|_2^2+\beta_{1t} \|\boldsymbol{v}_t^{1/4}(\boldsymbol{\theta}_{t-1}-\boldsymbol{\theta}^{\ast})\|_2^2,
\label{equ:convex1}
\end{align}
where the above inequality uses the Cauchy-Schwartz inequality $2\langle\boldsymbol{a},\boldsymbol{b}\rangle\leq \|\boldsymbol{a}\|_2^2+\|\boldsymbol{b}\|_2^2$. Note that (\ref{equ:convex1}) corresponds to (2) in the appendix of \cite{Zhuang20Adabelief} for AdaBelief.  

Summing (\ref{equ:convex1}) from $t=1$ until $t=T$,  rearranging the quantities, and exploiting the property that $\boldsymbol{g}_t=\nabla f(\boldsymbol{\theta}_{t-1})$ and $f(\cdot)$ being convex gives 
\begin{align}
&\hspace{-40mm}f(\bar{\boldsymbol{\theta}}_T) - f(\boldsymbol{\theta}^{\ast}) \nonumber \\
&\hspace{-40mm}=f\left(\frac{1}{T}\sum_{t=0}^{T-1}\boldsymbol{\theta}_t\right) - f(\boldsymbol{\theta}^{\ast}) \nonumber \\
&\hspace{-40mm}\stackrel{(a)}{\leq} \frac{1}{T}\sum_{t=1}^{T}\left(f(\boldsymbol{\theta}_{t-1}) - f(\boldsymbol{\theta}^{\ast})\right) \nonumber\\
&\hspace{-40mm}\stackrel{(b)}{\leq} \frac{1}{T}\sum_{t=1}^{T}\langle \boldsymbol{g}_t, \boldsymbol{\theta}_{t-1}-\boldsymbol{\theta}^{\ast} \rangle%\;\;\;\;\;\;\;\;\;\;\;\;\;\;\;
%\;\;\;\;\;\;\;\;\;\;\;\;\;\;\;
%\;\;\;\;\;\;\;\;\;\;\;\;\;\;\;
%\;\;\;\;\;\;\;\;\;\;\;\;\;\;\;
%\;\;\;\;\;\;\;\;\;\;\;\;\;\;\;
%\;\;\;\;\;\;\;\;\;\;\;\;\;\;\;
%\;\;\;\;\;\;\;\;\;\;\;\;\;\;\;
\nonumber 
\end{align}
\begin{align}
&\leq \frac{1}{T}\sum_{t=1}^T\Big[ \frac{1}{2\eta_t(1-\beta_{1t})} \Big(\|\boldsymbol{v}_t^{1/4}(\boldsymbol{\theta}_{t-1}-\boldsymbol{\theta}^{\ast}) \|_2^2  \nonumber\\
&\hspace{35mm}-\|\boldsymbol{v}_t^{1/4}(\boldsymbol{\theta}_t-\boldsymbol{\theta}^{\ast}) \|_2^2 \Big) \nonumber \\
&\hspace{3mm} +\frac{\eta_t}{2(1\hspace{-0.6mm}-\hspace{-0.6mm}\beta_{1t})}\|\boldsymbol{v}_t^{-1/4}\boldsymbol{m}_t\|_2^2 \hspace{-0.6mm}+\hspace{-0.6mm} \frac{\eta_t\beta_{1t}}{2(1 \hspace{-0.6mm}-\hspace{-0.6mm}\beta_{1t})}\|\boldsymbol{v}_t^{-1/4}\boldsymbol{m}_{t-1}\|_2^2 \nonumber \\
&\hspace{3mm}+ \frac{\beta_{1t}}{2\eta_t(1-\beta_{1t})}\|\boldsymbol{v}_t^{1/4}(\boldsymbol{\theta}_{t-1}-\boldsymbol{\theta}^{\ast})\|_2^2 \Big] \nonumber \\
&\stackrel{\beta_{11}=\beta_1}{=}\frac{1}{2\eta(1-\beta_1)T}\|\boldsymbol{v}_1^{1/4}(\boldsymbol{\theta}_0-\boldsymbol{\theta}^{\ast})\|_2^2 \nonumber \\
&\hspace{5mm}+\frac{1}{T}\sum_{t=1}^{T-1}\Bigg(\frac{1}{2\eta_{t+1}(1-\beta_{1(t+1)})}\|\boldsymbol{v}_{t+1}^{1/4}(\boldsymbol{\theta}_t-\boldsymbol{\theta}^{\ast})\|_2^2 \nonumber \\
&\hspace{7mm}-\frac{1}{2\eta_t(1-\beta_{1t})}\|\boldsymbol{v}_t^{1/4}(\boldsymbol{\theta}_t-\boldsymbol{\theta}^{\ast})\|_2^2\Bigg) \nonumber \\
&\hspace{2mm} +\hspace{-0.7mm}\frac{1}{T}\sum_{t=1}^T\hspace{-2mm}\Big[\frac{\eta_t}{2(1\hspace{-0.7mm}-\hspace{-0.7mm}\beta_{1t})}\|\boldsymbol{v}_t^{-1/4}\boldsymbol{m}_t\|_2^2+\hspace{-0.6mm}\frac{\eta_t\beta_{1t}}{2(1\hspace{-0.7mm}-\hspace{-0.7mm}\beta_{1t})}\|\boldsymbol{v}_t^{-1/4}\boldsymbol{m}_{t-1}\|_2^2 \nonumber\\
&\hspace{20mm}+ \frac{\beta_{1t}}{2\eta_t(1-\beta_{1t})}\|\boldsymbol{v}_t^{1/4}(\boldsymbol{\theta}_{t-1}-\boldsymbol{\theta}^{\ast})\|_2^2 \Big] \nonumber \\
&\hspace{3mm}\textcolor{blue}{\left(\textrm{condition: }\left\{ \begin{array}{l}0\leq \boldsymbol{v}_{t-1}[i]\leq \boldsymbol{v}_t[i] \textrm{ for all } i=1,\ldots, d, \\ 0\leq \eta_t\leq \eta_{t-1}, 0\leq \beta_{1(t+1)}\leq \beta_{1t}<1\end{array}\right)\right.} \nonumber \\
&\leq \frac{1}{2\eta(1-\beta_1)T}\|\boldsymbol{v}_1^{1/4}(\boldsymbol{\theta}_0-\boldsymbol{\theta}^{\ast})\|_2^2 \nonumber\\
&\hspace{5mm}+\frac{1}{T}\sum_{t=1}^{T-1}\Bigg(\frac{1}{2\eta_{t+1}(1-\beta_{1t})}\|\boldsymbol{v}_{t+1}^{1/4}(\boldsymbol{\theta}_t-\boldsymbol{\theta}^{\ast})\|_2^2\nonumber \\
&\hspace{20mm}-\frac{1}{2\eta_{t+1}(1-\beta_{1t})}\|\boldsymbol{v}_{t+1}^{1/4}(\boldsymbol{\theta}_t-\boldsymbol{\theta}^{\ast})\|_2^2\Bigg) \nonumber \\
&\hspace{2mm} +\hspace{-0.7mm}\frac{1}{T}\hspace{-0.7mm}\sum_{t=1}^T\hspace{-0.7mm}\Big[\frac{\eta_t}{2(1\hspace{-0.7mm}-\hspace{-0.7mm}\beta_{1})}\|\boldsymbol{v}_t^{-1/4}\boldsymbol{m}_t\|_2^2\hspace{-0.7mm}+\hspace{-0.7mm}\frac{\eta_t\beta_{1}}{2(1\hspace{-0.7mm}-\hspace{-0.7mm}\beta_{1})}\|\boldsymbol{v}_{t-1}^{-1/4}\boldsymbol{m}_{t-1}\|_2^2 \nonumber\\
&\hspace{20mm}+ \frac{\beta_{1t}}{2\eta_t(1-\beta_{1})}\|\boldsymbol{v}_t^{1/4}(\boldsymbol{\theta}_{t-1}-\boldsymbol{\theta}^{\ast})\|_2^2 \Big] \nonumber \\
&\hspace{3mm}\textcolor{blue}{(\textrm{condition:} 0\leq \eta_t\leq \eta_{t-1}, \boldsymbol{m}_0=\boldsymbol{0})} \nonumber \\
&\leq \frac{1}{2\eta(1-\beta_1)T}\|\boldsymbol{v}_1^{1/4}(\boldsymbol{\theta}_0-\boldsymbol{\theta}^{\ast})\|_2^2 \nonumber\\
&\hspace{4.5mm}+\frac{1}{T}\sum_{t=1}^T\frac{\eta_t(1+\beta_1)}{2(1-\beta_{1})}\|\boldsymbol{v}_t^{-1/4}\boldsymbol{m}_t\|_2^2  \nonumber\\
&\hspace{5mm}+ \frac{1}{T(1-\beta_{1})}\sum_{t=1}^T \frac{\beta_{1t}}{2\eta_t}\|\boldsymbol{v}_t^{1/4}(\boldsymbol{\theta}_{t-1}-\boldsymbol{\theta}^{\ast})\|_2^2 \nonumber \\
&\hspace{3mm}\textcolor{blue}{(\textrm{condition: }\|\boldsymbol{\theta}^{\ast}\|_{\infty} \leq D,   \|\boldsymbol{\theta}_t\|_{\infty} \leq D)}  \nonumber \\
&\hspace{0mm}\leq \frac{D^2}{\eta(1\hspace{-0.7mm}-\hspace{-0.7mm}\beta_1)T}\sum_{i=1}^d (\boldsymbol{v}_1[i])^{1/2} \hspace{-0.7mm}+\hspace{-0.7mm}\frac{1}{T}\sum_{t=1}^T\frac{\eta_t(1\hspace{-0.7mm}+\hspace{-0.7mm}\beta_1)}{2(1\hspace{-0.7mm}-\hspace{-0.7mm}\beta_{1})}\|\boldsymbol{v}_t^{-1/4}\boldsymbol{m}_t\|_2^2  \nonumber\\
&\hspace{5mm}+ \frac{D^2}{T(1-\beta_{1})}\sum_{t=1}^T \frac{\beta_{1t}}{\eta_t}\sum_{i=1}^d(\boldsymbol{v}_t[i])^{1/2} \nonumber \\
&\hspace{0mm}\stackrel{(c)}{\leq} \frac{D^2}{\eta(1\hspace{-0.7mm}-\hspace{-0.7mm}\beta_1)T}\sum_{i=1}^d (\boldsymbol{v}_1[i])^{1/2} \hspace{-0.7mm}+\hspace{-0.7mm}\frac{1}{T}\sum_{t=1}^T\frac{\eta_t(1\hspace{-0.7mm}+\hspace{-0.7mm}\beta_1)}{2(1\hspace{-0.7mm}-\hspace{-0.7mm}\beta_{1})}\|\boldsymbol{v}_t^{-1/4}\boldsymbol{m}_t\|_2^2  \nonumber\\
&\hspace{5mm}+ \frac{D^2\beta_1G_{\infty}d}{T(1-\beta_{1})\eta (1-\lambda)^2},
\label{equ:convex2}
\end{align}
where both step~$(a)$ and $(b)$ use the property of $f(\cdot)$ being convex, and step~$(c)$ uses the following conditions 
\begin{align}
\textcolor{blue}{\left\{\hspace{-2mm}\begin{array}{l}\|\boldsymbol{g}_t\|_{\infty}\hspace{-0.7mm}\leq\hspace{-0.7mm} G_{\infty}/2\Rightarrow \|\boldsymbol{m}_{l,t}^{(K)} \hspace{-0.7mm}-\hspace{-0.7mm} \boldsymbol{g}_{l,t}^{(K)}\|_{\infty}\hspace{-0.7mm}\leq\hspace{-0.7mm} G_{\infty}  \Rightarrow \|\boldsymbol{v}_t\|_{\infty} \hspace{-0.7mm}\leq\hspace{-0.7mm} G_{\infty}^2 \nonumber\\
\sum_{t=1}^T \frac{\beta_{1t}}{\eta_t}\leq\frac{\beta_{1}}{\eta}\sum_{t=1}^T \lambda^{t-1}\sqrt{t}\leq \frac{\beta_{1}}{\eta}\sum_{t=1}^T \lambda^{t-1}t \leq \frac{\beta_1}{\eta(1-\lambda)^2}, \end{array}\right.}
\end{align}
which are obtained from the appendices of \cite{Zhuang20Adabelief}. 

Next we consider the quantity $\sum_{t=1}^T\eta_t \|\boldsymbol{v}_t^{-1/4}\boldsymbol{m}_t\|_2^2 $ in (\ref{equ:convex2}), the upper bound of which is given in \cite{Zhuang20Adabelief}: 
\begin{lemma}[Equ.~(4) in the appendix of \cite{Zhuang20Adabelief}] Let ${\boldsymbol{g}_{1:T}^2[i]}=((\boldsymbol{g}_1[i])^2,\ldots, (\boldsymbol{g}_T[i])^2)\in \mathbb{R}^{T}$. Under the three assumptions given in the theorem, we have 
\begin{align}
    \sum_{t=1}^T\eta_t\|\boldsymbol{v}_t^{-1/4}\boldsymbol{m}_t\|_2^2  \leq \frac{\eta\sqrt{1+\log T}}{\sqrt{c}(1-\beta_1)^2}\|(\boldsymbol{g}_{1:T}^2[i])\|_2. 
    \label{equ:convex3}
\end{align}
\end{lemma}

Finally, plugging (\ref{equ:convex3}) into (\ref{equ:convex2}) produces the upper-bound regret in the theorem. The proof is complete.   
\end{proof}

%The obtained regret bound is tighter than the one derived in \cite{Zhuang20Adabelief} for AdaBelief due to improved mathematical derivation. In particular, the first term in Theorem~\ref{theorem:convex} is of order $O(1/T)$ while the corresponding one in \cite{Zhuang20Adabelief} is essentially of order $O(1/\sqrt{T})$. In other words, we improved the regret bound of \cite{Zhuang20Adabelief}.
%The above theorem implies that the regret of Aida is upper bounded by $\mathcal{O}(1/\sqrt{T})$, which is consistent with the results of \cite{Zhuang20Adabelief} for AdaBelief. Similar regret bounds can also be obtained for LAdam and LAdabelief by using the assumption $0<c\leq q_{l,t-1} \leq q_{l,t}$ for all $l\in [L] $ and $t\in [T]$.  

\begin{table}[t!]
\caption{Performance comparison for
training the transformer.}
\label{tab:transformer_val_acc}
\centering
  \begin{tabular}{|c|c||c|c|}
  \cline{1-2}
    {\footnotesize $\begin{array}{c}\textrm{SGD}\\ \textrm{(non-adaptive)}\end{array}$}& \footnotesize{55.58$\pm$0.34}

    \\ 
  \hline
  {\footnotesize AdaBound} & \footnotesize{55.90$\pm$0.21}
  & \footnotesize{Yogi} & \footnotesize{60.47$\pm$0.61}  \\ \hline
  \footnotesize{RAdam} & \footnotesize{64.47$\pm$0.19} &
   \footnotesize{MSVAG} & \footnotesize{53.79$\pm$0.13} \\ \hline 
      \footnotesize{Fromage} & \footnotesize{35.57$\pm$0.19} & 
   \footnotesize{Adam} & \footnotesize{64.71$\pm$0.57} \\    \hline 
       \footnotesize{AdamW} & \footnotesize{64.49$\pm$0.24} &  
    \footnotesize{AdaBelief} & \footnotesize{66.90$\pm$0.77} \\
   \hline
  \footnotesize{Aida(K=1)} & \footnotesize{68.77$\pm$0.16} &
  \footnotesize{Aida(K=2)} & \footnotesize{\textbf{68.96}$\pm$0.06} \\ \hline 
  \end{tabular}
\end{table}

\begin{figure}[t!]
\centering
\includegraphics[width=80mm]{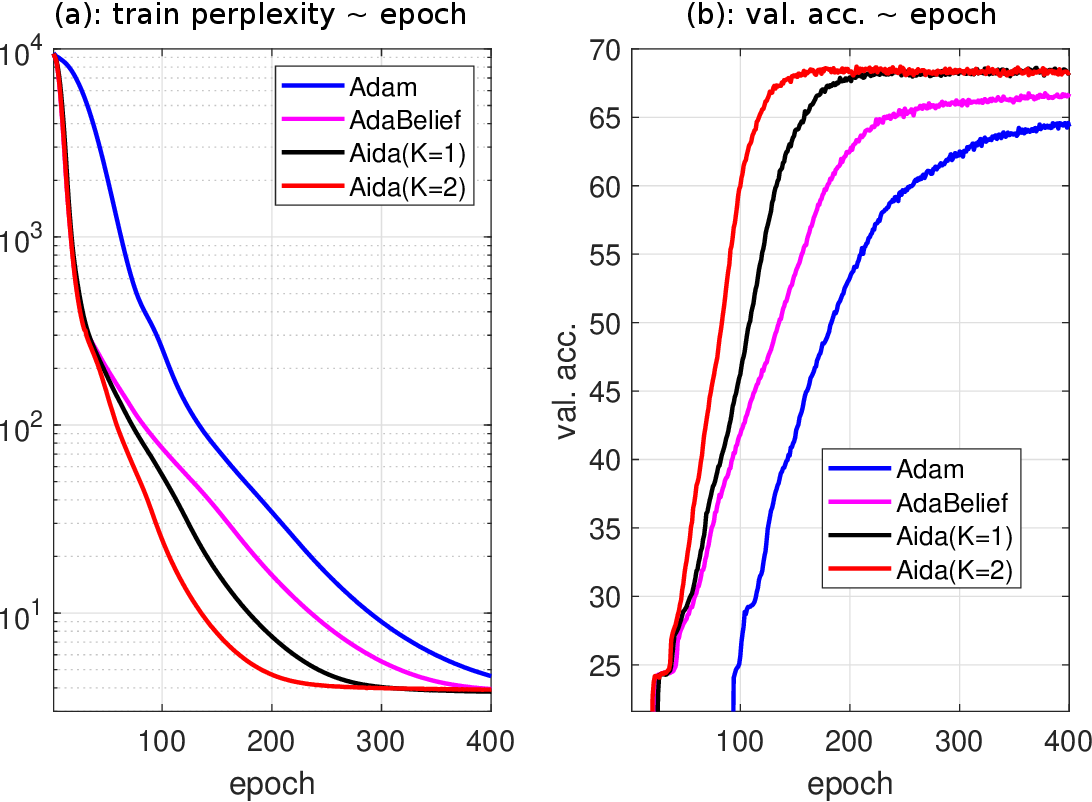}
\vspace*{-0.0cm}
\caption{\footnotesize{Performance visualisation of Aida, Adam, and AdaBelief for the training of the transformer. } } 
\label{fig:trans_comapre}
\vspace*{-0.2cm}
\end{figure}

\vspace{-2mm}
\section{Experiments}
\label{sec:exp}
\vspace{-1mm}

\begin{table*}[t!]
\caption{\small  Validation perplexity on Penn Treebank for 1, 2, 3-layer LSTM. \textbf{lower} is better.  \vspace{-2mm} } 
\label{tab:LSTM}
\centering
\begin{tabular}{|c|c|c|c|c|c|c|}
\hline
 &   {{\footnotesize Aida($K$=1) }}
& \hspace{0mm} {{\footnotesize AdaBelief  }}\hspace{0mm}
& \hspace{0mm}{\footnotesize  AdamW }   \hspace{0mm} & {\footnotesize Adam}
 & {\footnotesize Yogi}  & \footnotesize{AdaBound} \\
\hline 
\footnotesize{1 layer}&  
\footnotesize{ 82.27} &
\footnotesize{ 84.21} &
\footnotesize{88.36}    & \footnotesize{ 84.28 } & \footnotesize{86.78} & \footnotesize{84.52}
\\ \hline 
\footnotesize{2 layer}& 
\footnotesize{66.16} &
\footnotesize{66.29} &
\footnotesize{73.18}   & \footnotesize{66.86} & \footnotesize{71.56}  & \footnotesize{67.01}
\\ \hline 
\footnotesize{3 layer}& 
\footnotesize{61.98} &
\footnotesize{61.23} &
\footnotesize{70.08}   & \footnotesize{64.28} & \footnotesize{67.83}  & \footnotesize{63.16} \\ 
\hline
%\hline % inserts single-line
\cline{1-6}
& \footnotesize{Aida(K=2)}  & {\footnotesize $\begin{array}{c}\textrm{SGD}\\ \textrm{(non-adaptive)}\end{array}$} & {\footnotesize RAdam}
 & {\footnotesize MSVAG} & {Fromage}
 \\
\cline{1-6}
\footnotesize{1 layer}& \footnotesize{\textbf{81.53}} 
& \footnotesize{85.52}
 & \footnotesize{88.76}
 & \footnotesize{84.75}
 & \footnotesize{85.20}
 \\
\cline{1-6}
\footnotesize{2 layer}& \footnotesize{\textbf{65.04}}
& \footnotesize{67.44}
 & \footnotesize{74.12}
 & \footnotesize{ 68.91}
 & \footnotesize{ 72.22} 
 \\
\cline{1-6}
\footnotesize{3 layer}& \footnotesize{\textbf{60.18}}
& \footnotesize{63.68}
 & \footnotesize{70.41}
 & \footnotesize{65.04}
 & \footnotesize{67.37}
 \\
\cline{1-6}
\end{tabular}
\vspace{-1mm}
\end{table*}

\begin{figure*}[t!]
\centering
\includegraphics[width=120mm]{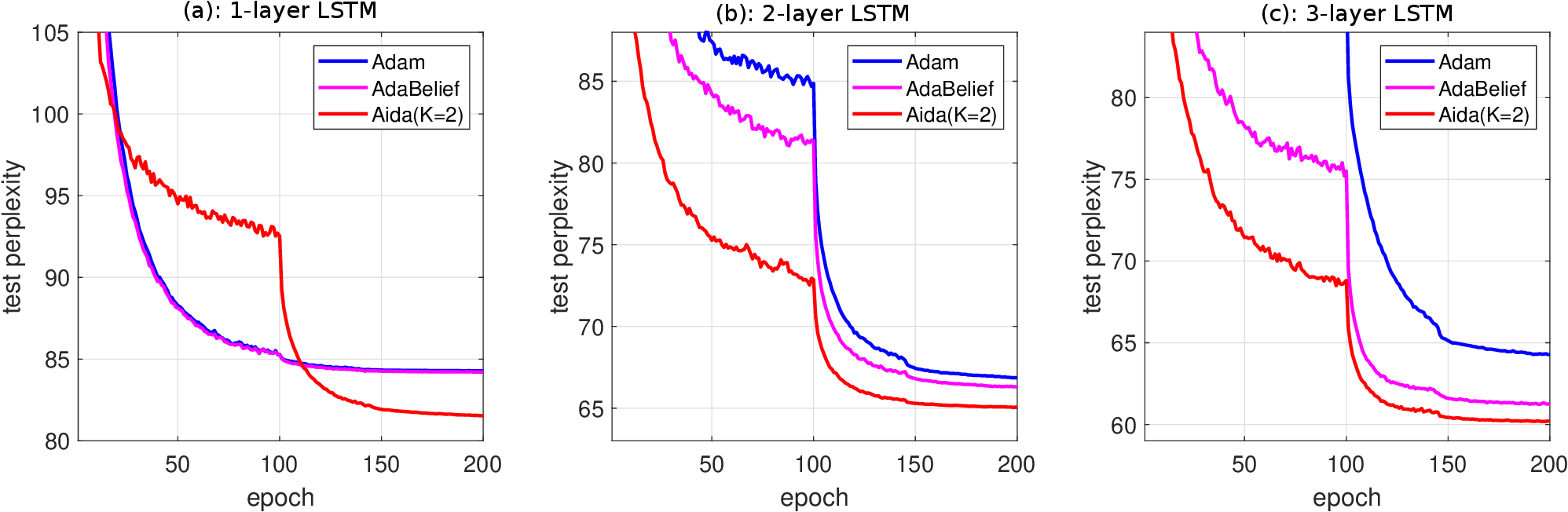}
\vspace*{-0.0cm}
\caption{\footnotesize{Performance visualisation of Aida, AdaBelief, and Adam in Table~\ref{tab:LSTM}. } }
\label{fig:LSTM_comapre}
\vspace*{-0.3cm}
\end{figure*}

We evaluated Aida on three types of DNN tasks: (1) natural language processing (NLP) on training transformer and LSTM models; (2) image classification on training VGG and ResNet \cite{He15ResNet} models; (3) image generation on training WGAN-GP \cite{Gulrajani17WGANGP}.  Two open-source repositories\footnote{
``https://github.com/jadore801120/attention-is-all-you-need-pytorch" is adopted for the task of training a transformer, which produces reasonable validation performance using Adam.
``https://github.com/juntang-zhuang/Adabelief-Optimizer" is adopted for all the remaining tasks. The second open source is the original implementation of AdaBelief \cite{Zhuang20Adabelief}. } were used for the above DNN training tasks. To demonstrate the effectiveness of the proposed method, eight adaptive optimization algorithms from the literature were tested and compared, namely Yogi  \cite{Zaheer18Yogi}, RAdam \cite{Liu19RAdam}, MSVAG  \cite{Balles17MSVAG}, Fromage \cite{Bernstein20Fromage}, Adam \cite{Kingma17}, AdaBound \cite{Luo19AdaBound},  AdamW \cite{Loshchilov19AdamW}, and AdaBelief \cite{Zhuang20Adabelief}. In addition, SGD with momentum was evaluated as a baseline for performance comparison. In all experiments, the additional parameter $\xi$ in Algorithm~1 was set to $\xi=1e-20$.

It is found that Aida with $K=2$ outperforms the nine reference methods for training transformer, LSTM, VGG11, and ResNet34 models while it matches the best performance of the nine methods for training WGAN-GP. Furthermore, experiments on training ResNet18 on the large ImageNet dataset show that Aida outperforms AdaBelief. 

%either better or competitive in comparison to a number of popular adaptive optimization methods.  

The time complexity of Aida was evaluated for training VGG11 and ResNet34 on a 2080 Ti GPU. In brief, Aida with $K=2$ consumed $25\%$ more time per epoch compared to AdaBelief.

\vspace{-1mm}
\subsection{On training a transformer}
 In this task, we consider the training of a transformer for WMT16: multimodal translation by using the first open-source as indicated in the footnote.  In the training process, we retained almost all of the default hyper-parameters provided in the open-source except for the batch size. Due to limited GPU memory, we changed the batch size from 256 to 200.  
The parameters of Aida were set to  $(\eta_0,\beta_1, \beta_2, \epsilon)=(0.001, 0.9, 0.98, 1e-16)$. The parameter-setups for other optimizers can be found in Table~\ref{tab:setup_transformer} of Appendix~\ref{appendix:parameterSetups}, where certain hyper-parameters for each optimizer were searched over some discrete sets to optimize the validation performance. For example, the parameter $\epsilon$ of Adam was searched over the set $\{1e-6,1e-7,\ldots, 1e-12\}$ while the remaining parameters were set to $(\eta_0,\beta_1,\beta_2)=(0.001, 0.9, 0.98)$ as in Aida.  Once the optimal parameter-configuration for each optimizer was obtained by searching, three experimental repetitions were then performed to alleviate the effect of the randomness. 

It is clear from Table~\ref{tab:transformer_val_acc} and Fig.~\ref{fig:trans_comapre} that Aida significantly outperforms all other methods. We emphasize that the maximum number of epochs was set to 400 for each optimizer by following the default setup in the open source, and no epoch cutoff is performed in favor of Aida.  When $K$ increases from 1 to 2 in Aida, the method converges considerably faster and produces better validation performance, which may be due to the fact that Aida with $K=2$ has a more compact range of adaptive stepsizes. On the other hand, the non-adaptive method SGD with momentum produces a performance that is inferior to all adaptive methods except Fromage and MSVAG.

%{
%  \caption{Validation accuracies for training a transformer}%
%}
%\begin{table}[h!]
%\caption{\small  Validation perplexities for training a Transformer  \vspace{2mm} } 
%\label{tab:transformer_val_acc}
%\centering
%\begin{tabular}{|c|c|c|c|c|}
%\hline
% \hspace{0mm} {{\scriptsize Aida($K$=2)}}\hspace{0mm}
%& \hspace{0mm} {{\scriptsize Aida($K$=1)}}\hspace{0mm}
%& \hspace{0mm}{\scriptsize  AdaBelief }   \hspace{0mm} & {\scriptsize Adam}
% & {\scriptsize Yogi}  \\
%\hline 
%\scriptsize{\textbf{68.96} $\pm$ 0.06} & \scriptsize{68.77 $\pm$ 0.16} &\scriptsize{ 66.90 $\pm$ 0.77  }   & \scriptsize{ 64.71 $\pm$ 0.57 } & \scriptsize{ 60.47 $\pm$ XX } \\ 
%\hline
%\hline  % inserts single-line
%\scriptsize{LAdam} & \scriptsize{LAdaBelief}  & \scriptsize{SGD} & {\scriptsize RAdam}
% & {\scriptsize MSVAG}
% \\
%\hline
%\scriptsize{66.66$\pm$0.26} &
%\scriptsize{66.35$\pm$ 0.96}
%& \scriptsize{55.58$\pm$ 0.34 }
% & \scriptsize{ 64.54 $\pm$ XX }
% & \scriptsize{ 53.50 $\pm$ XX }
%\\ % Entering row contents Midnight&7&-3& 5&3&-1&-3&5\\
%\hline
%\end{tabular}
%\vspace{-1mm}
%\end{table}

\begin{table*}[t]
\caption{\footnotesize Validation accuracies (in percentage) and \textcolor{blue}{time complexity in seconds per epoch (referred to as \emph{t. c.})} of nine methods for training VGG11 and ResNet34 over CIFAR10 and CIFAR100. The best result is highlighted in each column. \vspace{-2mm}
} 
\label{tab:val_acc_imgclass}
\centering
%\begin{tabular}{ |*{8}{c|} } 
\begin{tabular}{|c|c|c|c|c||c|c|c|c|}
\cline{2-9} 
%\hline
%{\scriptsize \multicolumn{9}{c}{deterministic} }
\multicolumn{1}{c|}{} & \multicolumn{4}{|c||}{{\footnotesize CIFAR10 }} & \multicolumn{4}{|c|}{{\footnotesize CIFAR100 }}      \\
\cline{2-9} %\cline{1-3}
 \multicolumn{1}{c|}{} & \multicolumn{2}{|c|}{\footnotesize{VGG11}} & \multicolumn{2}{|c||}{\footnotesize{ResNet34}} & \multicolumn{2}{|c|}{\footnotesize{VGG11}} & \multicolumn{2}{|c|}{\footnotesize{ResNet34}} %& \scriptsize{X}& \scriptsize{X} 
 \\ % Entering row contents Midnight&7&-3& 5&3&-1&-3&5\\
 \hline %\cline{1-3} %\cline{7-8}
 \multicolumn{1}{|c|}{{\footnotesize optimizers}} & \footnotesize{val. acc} & \footnotesize{t.~c.} & \footnotesize{val. acc} & \footnotesize{t.~c.} &
 \footnotesize{val. acc} & \footnotesize{t.~c.} &
 \footnotesize{val. acc} & \footnotesize{t.~c.} 
 \\
 \hline
{\footnotesize $\begin{array}{c}\textrm{SGD}\\ \textrm{(non-adaptive)}\end{array}$} \hspace{-2mm} & \hspace{-2mm} \footnotesize{91.36$\pm$0.07} & \footnotesize{\textbf{5.83}} \hspace{-2mm} & \hspace{-2mm} {\footnotesize{{95.48}$\pm$0.11}} & \footnotesize{\textbf{30.45}} \hspace{-2mm} & \footnotesize{ 67.02$\pm$0.25} & \footnotesize{\textbf{5.85}} &  {\footnotesize{{78.10}$\pm$0.18}} & \footnotesize{\textbf{30.92}} %& \hspace{-2mm} \scriptsize{X} \hspace{-2mm} & \hspace{-2mm} \scriptsize{X}  \hspace{-2mm} 
 \\ 
\hline
\hspace{-2mm} {\footnotesize Yogi} \hspace{-2mm} & \hspace{-2mm} \footnotesize{90.74$\pm$0.16} & \footnotesize{6.49}  \hspace{-2mm} & \hspace{-2mm} \footnotesize{94.98$\pm$0.26}  & \footnotesize{31.74}  \hspace{-2mm} &  \footnotesize{65.57$\pm$0.17} & \footnotesize{6.42} &  \footnotesize{77.17}$\pm$0.12 & \footnotesize{32.20} %& \hspace{-2mm} \scriptsize{X} \hspace{-2mm} & \hspace{-2mm} \scriptsize{X}  \hspace{-2mm} 
\\ % Entering row contents Midnight&7&-3& 5&3&-1&-3&5\\
\hline %\cline{1-3} % \cline{7-8}
{\footnotesize RAdam} \hspace{-2mm} & \hspace{-2mm} \footnotesize{89.58$\pm$}0.10 & \footnotesize{6.28} \hspace{-2mm} & \hspace{-2mm} \footnotesize{94.64$\pm$0.18} & \footnotesize{31.21} \hspace{-2mm} & \footnotesize{63.62$\pm$0.20} & \footnotesize{6.29} & \footnotesize{74.87}$\pm$0.13 & \footnotesize{31.58} %& \hspace{-2mm} \scriptsize{X} \hspace{-2mm} & \hspace{-2mm} \scriptsize{X}  \hspace{-2mm}  
\\
\hline %\cline{1-3} % \cline{7-8}
{\footnotesize MSVAG} \hspace{-2mm} & \hspace{-2mm} \footnotesize{90.04$\pm$0.22} & \footnotesize{7.08} \hspace{-2mm} & \hspace{-2mm} \footnotesize{94.65$\pm$0.08} & \footnotesize{33.78} \hspace{-2mm} & \footnotesize{62.67$\pm$0.33} & \footnotesize{7.19} & \footnotesize{75.57}$\pm$0.14
& \footnotesize{33.80}
%& \hspace{-2mm} \scriptsize{X} \hspace{-2mm} & \hspace{-2mm} \scriptsize{X}  \hspace{-2mm}  
\\
 \hline %\cline{1-3} %\cline{7-8}
{\footnotesize Fromage} \hspace{-2mm} & \hspace{-2mm} \footnotesize{89.72$\pm$0.25} & \footnotesize{6.66} \hspace{-2mm} & \hspace{-2mm}  \footnotesize{94.64$\pm$0.07} & \footnotesize{35.19} \hspace{-2mm} & \footnotesize{62.93$\pm$0.53} & \footnotesize{6.56} & \footnotesize{74.84}$\pm$0.27 & \footnotesize{35.50} %& \hspace{-2mm} \scriptsize{} \hspace{-2mm} & \hspace{-2mm} \scriptsize{X}  \hspace{-2mm}  
\\
 \hline %\cline{1-3} %\cline{7-8}
{\footnotesize Adam} \hspace{-2mm} & \hspace{-2mm} \footnotesize{91.20$\pm$0.21} & \footnotesize{6.15} \hspace{-2mm} & \hspace{-2mm} \footnotesize{95.09$\pm$0.18} & \footnotesize{31.28} \hspace{-2mm} & \footnotesize{67.88$\pm$0.13} & \footnotesize{6.20} & \footnotesize{77.31}$\pm$0.14 & \footnotesize{31.47} %& \hspace{-2mm} \scriptsize{X} \hspace{-2mm} &d  \hspace{-2mm} \scriptsize{X}  \hspace{-2mm}  
\\
 \hline %\cline{1-3} % \cline{7-8}
{\footnotesize AdamW} \hspace{-2mm} & \hspace{-2mm}   \footnotesize{89.46$\pm$0.08} & \footnotesize{6.25} \hspace{-2mm} & \hspace{-2mm} \footnotesize{94.48$\pm$0.18} & \footnotesize{31.71} \hspace{-2mm} & \footnotesize{62.50$\pm$0.23} & \footnotesize{6.31} & \footnotesize{74.29$\pm$0.20} & \footnotesize{31.80} %& \hspace{-2mm} \scriptsize{X} \hspace{-2mm} & \hspace{-2mm} \scriptsize{X}  \hspace{-2mm}  
\\
 \hline %\cline{1-3} % \cline{7-8}
 {\footnotesize AdaBound}  \hspace{-2mm} & \hspace{-2mm} {\footnotesize{90.48$\pm$0.12}} & \footnotesize{6.71}  \hspace{-2mm} & \hspace{-2mm}  \footnotesize{94.73$\pm$0.16} & \footnotesize{33.75} \hspace{-2mm} & {\footnotesize{64.80$\pm$0.42}} %epsilon=1e-9
 & \footnotesize{6.73} & \footnotesize{76.15}$\pm$0.10 & \footnotesize{33.78}  %& \hspace{-2mm} \scriptsize{92.29} \hspace{-2mm} & \hspace{-2mm} \scriptsize{92.53}  \hspace{-2mm}  
\\
\hline
 {\footnotesize AdaBelief}  \hspace{-2mm} & \hspace{-2mm} {\footnotesize{91.55$\pm$0.13}} & \footnotesize{6.47}  \hspace{-2mm} & \hspace{-2mm}  \footnotesize{95.15$\pm$0.11} & \footnotesize{31.66} \hspace{-2mm} & {\footnotesize{68.05$\pm$0.31}} %epsilon=1e-9
 & \footnotesize{6.49} & \footnotesize{77.32}$\pm$0.37 & \footnotesize{31.74}  %& \hspace{-2mm} \scriptsize{92.29} \hspace{-2mm} & \hspace{-2mm} \scriptsize{92.53}  \hspace{-2mm}  
\\
\hline
\hline%{\arrayrulewidth}
%\cline{1-3}  %\cline{7-8}
{\footnotesize Aida($K$=1) } \hspace{-2mm} & \hspace{-2mm} {\footnotesize{91.52$\pm$0.05}}  & \footnotesize{7.27} \hspace{-2mm} & \hspace{-2mm}  {\footnotesize{95.31$\pm$0.05}}  & \footnotesize{35.25}  \hspace{-2mm} & {\footnotesize{{68.89}$\pm$0.09}} & \footnotesize{7.32} & {\footnotesize{77.50}$\pm$0.12} & \footnotesize{35.46}
\\ 
\hline
 {\footnotesize Aida($K$=2)} \hspace{-2mm} & \hspace{-2mm} {\footnotesize{\textbf{91.68}$\pm$0.16}} & \footnotesize{7.95} \hspace{-2mm} & \hspace{-2mm} {\footnotesize{\textbf{95.57}$\pm$0.13}} & \footnotesize{39.53}  \hspace{-2mm}  & {\footnotesize{\textbf{69.02}$\pm$0.11}} & \footnotesize{8.01} & {\footnotesize{\textbf{78.86}}$\pm$0.12} & \footnotesize{39.64}
 \\ \hline
\end{tabular}
\vspace{-0.5mm}
\end{table*}

\begin{table*}[h!]
\caption{\small Best FID obtained for each optimizer (\textbf{lower}
is better)   \vspace{-2mm} } 
\label{tab:GAN_FID}
\centering
\begin{tabular}{|c|c|c|c|c|c|c|}
\hline
& \hspace{0mm} {{\footnotesize Aida($K$=2) }}\hspace{0mm}
& \hspace{0mm}{\footnotesize  Aida($K$=1) }   \hspace{0mm} & {\footnotesize AdaBelief}
& {\footnotesize Adam} &  {\footnotesize RAdam}
& {\footnotesize AdaBound}
\\ \hline
\footnotesize{best FIDs} & \footnotesize{55.7} & \footnotesize{\textbf{55.65}}  & \footnotesize{56.73} & 66.71 & \footnotesize{69.14} & \footnotesize{61.65}   \\
\hline 
\cline{1-6}

&  {\footnotesize AdamW} & {\footnotesize MSVAG} & {\footnotesize SGD} & {\footnotesize Yogi} & {\footnotesize Fromage}  \\
\cline{1-6}
\footnotesize{best FIDs} &  \footnotesize{63.76} &  \footnotesize{69.47} & \footnotesize{90.61} & \footnotesize{68.34} &  \footnotesize{78.47}
\\
\cline{1-6}
\end{tabular}
\end{table*}

\subsection{On training LSTMs}
In this experiment, we consider training LSTMs with a different number of layers over the Penn TreeBank dataset \cite{Marcus93PennTree}. The detailed experimental setup such as dropout rate and gradient-clipping magnitude can be found in the first open-source repository provided in the footnote. The parameters of Aida were set to $(\eta_0, \beta_1,\beta_2,\epsilon)=(0.001,0.9,0.999,1e-16)$. Similar to the task of training the transformer, the other optimizers have both fixed and free parameters of which the free parameters remain to be searched over some discrete sets. See Table~\ref{tab:setup_LSTM} in Appendix~\ref{appendix:parameterSetups} for a summary of the fixed and free parameters for each optimizer. An example is Adam for which $\eta_0\in \{0.01, 0.001\}$ and $\epsilon\in \{1e-6, 1e-8, 1e-10, 1e-12\}$ were tested to find the optimal configuration that produces the best validation performance.   
  
Table.~\ref{tab:LSTM} summarises the obtained validation perplexities of the ten methods for training 1, 2, and 3-layer LSTMs. It was found that for each optimizer, independent experimental repetitions lead to almost the same validation perplexity value. Therefore, we only keep the average of the validation perplexity values from three independent experimental repetitions for each experimental setup in the table and ignore the standard deviations.

It is clear from Table.~\ref{tab:LSTM} that Aida again outperforms all other methods in all three scenarios, which may be due to the contribution of a compact range of adaptive stepsizes in Aida. Fig.~\ref{fig:LSTM_comapre} further visualised the validation performance of Aida compared to AdaBelief and Adam. The performance gain of Aida is considerable in all three scenarios.   %AdaBelief performs slightly better than LAdam and LAdaBelief. This might be because, on Pytorch platform \cite{NEURIPS2019_9015}, the different small weight matrices of a single LSTM cell are put into a big matrix, of which all the parameters are treated to be from a single neural layer. It may occur that those small weight matrices have different gradient statistics as each of them process different information within the LSTM cell.    

\vspace{-1mm}
\subsection{On training VGG11 and ResNet34 over CIFAR10 and CIFAR100}
In this task, the ten optimizers were evaluated by following  a similar experimental setup as in \cite{Zhuang20Adabelief}. In particular, the batch size and epoch were set to 128 and 200, respectively. The common stepsize $\eta_t$ is reduced by multiplying by 0.1 at 100 and 160 epoch.  The detailed parameter-setups for the optimizers can be found in Table~\ref{tab:setup_VGGResNet} in Appendix~\ref{appendix:parameterSetups}.  Three experimental repetitions were conducted for each optimizer to alleviate the effect of randomness.     

Both the validation performance and the algorithmic complexity are summarised in Table~\ref{tab:val_acc_imgclass}. It is clear that Aida with $K=2$ consistently outperforms the nine reference methods in terms of validation accuracies at the cost of additional computational time. This demonstrates that the compact range of adaptive stepsizes in Aida does indeed improve the generalization performance.

We can also conclude from the table that SGD with momentum is the most computationally efficient method. On the other hand, due to the layerwise vector projections, Aida with $K=2$ consumed an additional $25\%$ time per epoch compared to AdaBelief.

\vspace{-2mm}
\subsection{On training WGAN-GP over CIFAR10}
\vspace{-1mm}
This task focuses on training WGAN-GP. The parameters of Aida and AdaBelief were set to $(\eta_t, \beta_1,\beta_2, \epsilon)=(0.0002, 0.5, 0.999, 1e-12)$.
The other eight optimizers have both fixed and free parameters, details of which can be found in Table~\ref{tab:setup_WGANGP} of Appendix~\ref{appendix:parameterSetups}. As an example, the free parameter of Adam is $\epsilon\in\{1e-4, 1e-6,\ldots, 1e-14\}$. 
For each parameter-configuration of an optimizer, three experimental repetitions were performed due to the relatively unstable Frechet inception distance (FID) scores in training WGAN-GP. 

Table~\ref{tab:GAN_FID} shows the best FID for each method. Considering Adam for example, it has six parameter-configuration due to six $\epsilon$ values being tested. As a result, the best FID for Adam is obtained over 18 values, accounting for three experimental repetitions for each of six $\epsilon$ values. It can be seen from the table that Aida with $K\in \{1,2\}$ provides bettter performance than AdaBelief, while the other methods perform significantly worse.

\vspace{-1mm}
\subsection{On training ResNet18 over ImageNet} 
In the last experiment, we investigated the performance gain of Aida compared to AdaBelief for training ResNet18 on the large ImageNet dataset. The maximum epoch and minibatch size were set to 90 and 256, respectively. The common stepsize $\eta_t$ is dropped by a factor of 0.1 at 70 and 80 epochs.  The parameter setup for the two optimizers can be found in Table~\ref{tab:setup_imagenet} of Appendix~\ref{appendix:parameterSetups}. Similarly, three experimental repetitions were conducted for each optimizer to mitigate the effect of randomness. 

It is clear from Table~\ref{tab:imagenet} that for the large ImageNet dataset, Aida again performs better than AdaBelief, indicating that the performance gain of Aida is robust against different sizes of datasets. We note that due to prohibitive computational effort, Aida is only compared to the most relevant optimizer: AdaBelief in Table~\ref{tab:imagenet}. In \cite{Zhuang20Adabelief}, AdaBelief is found to outperform the other eight optimizers when training ResNet18 over ImageNet (see Table~2 of \cite{Zhuang20Adabelief} where no std is reported).

\begin{table}[h!]
\caption{\footnotesize Validation accuracies (in percentage) of AdaBelief and Aida for training ResNet18 over ImageNet.    \vspace{-2mm} } 
\label{tab:imagenet}
\centering
\begin{tabular}{|c|c|c|}
\hline
 {\footnotesize optimizers} & \hspace{0mm} {{\footnotesize AdaBelief }}\hspace{0mm}
&   \hspace{0mm} {{\footnotesize Aida($K=2$) }}\hspace{0mm}  
\\ \hline
\footnotesize{val. acc.}  & \footnotesize{69.65}$\pm$0.06  & \footnotesize{\textbf{69.70}}$\pm$0.08
\\ \hline
\end{tabular}
\end{table}

\vspace{-2mm}
\section{Conclusions}
\vspace{-1mm}

%In this paper, we have made three contributions. Firstly, we study the particular placement of $\epsilon$ in the update expressions of AdaBelief. We show that the inclusion of $\epsilon$ in the computation of the 2nd momentum $\boldsymbol{s}_t$ (see Eq.~(\ref{equ:AdaBelief})) in AdaBelief helps to reduce the range of adaptive stepsizes, making it closer to SGD with momentum. The above technique can also be applied to other Adam-like methods to reduce their range of adaptive stepsizes (e.g., see Adam+ in Appendix~\ref{appendix:adamplus}).  

In this paper, we have shown that the range of the adaptive stepsizes of DNN optimizers has a significant impact on performance. The proposed Aida optimizer suppresses the range of the adaptive stepsizes of AdaBelief making it closer to SGD with momentum. Our experimental results indicate that Aida will be able to produce better performance across a wide range of DNN-based applications.

In the design of the Aida optimizer, we  track the EMA (or equivalently the second momentum) of $(\gamma_{l,t}\boldsymbol{m}_{l,t}-\beta_{l,t}\boldsymbol{g}_{l,t})^2$ for the $l$th layer of a DNN model as opposed to $(\boldsymbol{m}_{l,t}-\boldsymbol{g}_{l,t})^2$ used in AdaBelief, where $\gamma_{l,t}, \beta_{l,t}\in (0,1]$ are obtained by vector projections. Consequently, the adaptive stepsizes of Aida have a more compact range than those of AdaBelief. 

Our empirical study shows that Aida with $K=2$ outperforms nine optimizers including Adam and AdaBelief for training transformer, LSTM, VGG11, and ResNet34 models while at the same time it matches the best performance of the nine methods for training WGAN-GP models. In addition, experiments on training ResNet18 over the large  ImageNet dataset show that Aida performs better than AdaBelief.  On the other hand, it was found that the non-adaptive method SGD with momentum only produces good performance when training VGG and ResNet models. This suggests that the \emph{adaptivity} of Aida is important, allowing the method to effectively train different types of DNN models.

\vspace{-2mm}
%\bibliographystyle{IEEEtran}
%\bibliography{sigProcessing}

% or


\appendices
%\clearpage
%\newpage 
%\onecolumn
\hspace{-2mm}
\section{Update procedure of AdaBelief*}
\label{appendix:AdaBeliefstar}
The first $\epsilon$ is removed in (\ref{equ:AdaBelief}) to verify if AdaBelief* has a broad range of adaptive stepsizes than AdaBelief. 
\label{appendix:LAdaBelief}
\begin{algorithm}[h!]
   \caption{\small AdaBelief*}
   \label{alg:Adam}
\begin{algorithmic}[1]
   \STATE {\small {\bfseries Input:} $\beta_1$, $\beta_2$,  $\eta_t$, $\epsilon > 0$ }
   \STATE {\small {\bfseries  Init.:} $\boldsymbol{\theta}_0\hspace{-0.5mm}\in\hspace{-0.5mm} \mathbb{R}^d$,  $\boldsymbol{m}_0 \hspace{-0.5mm}=\hspace{-0.5mm} 0$, $\boldsymbol{v}_{0}=0 \in \mathbb{R}^d$ }
   \FOR{\small $t=1, 2, \ldots, T$}
   \STATE \hspace{-0mm}{\small  $\boldsymbol{g}_t \leftarrow \nabla f({\boldsymbol{\theta}}_{t-1}) $  }
   \STATE \hspace{-0mm}{\small $\boldsymbol{m}_{t} \leftarrow \beta_1 \boldsymbol{m}_{t-1}  + (1-\beta_1) \boldsymbol{g}_t$ }
    \STATE \hspace{-0mm}{\small ${q}_{t}  \leftarrow \beta_2 q_{t-1} + (1-\beta_2) (\boldsymbol{m}_{t}-\boldsymbol{g}_{t})^2$}
  % \STATE {\small $  {x}_t \leftarrow {x}_{t-1} \hspace{-0.6mm} - \hspace{-0.6mm}  \left\{\hspace{-1.5mm} \begin{array}{l}  \eta \frac{ 1}{(1-\beta_1^t)^q}  \frac{|{m}_t|^{q-1}\odot m_t}{(r_t/(1-\beta_2^t))^{q/p}+ \epsilon}    \\
  %\eta \frac{ (1-\beta_2^t))^{q/p} }{(1-\beta_1^t)^q}  \frac{|{m}_t|^{q-1}\odot m_t}{(r_t)^{q/p}+ \epsilon} 
    %\end{array} \right. $ }
   \STATE \hspace{-0mm}{\small $  \tilde{\boldsymbol{m}}_{t} \hspace{-0.6mm}\leftarrow \frac{\boldsymbol{m}_{t}}{1-\beta_1^{t}}\quad   \tilde{q}_{t} \leftarrow \frac{q_{t}}{1-\beta_2^{t}} $ }     
  \STATE {\small  $ \boldsymbol{\theta}_{t} \hspace{-0.6mm}\leftarrow \boldsymbol{\theta}_{t-1} -\frac{\eta_t }{\sqrt{\tilde{q}_{t}}+\epsilon} \tilde{\boldsymbol{m}}_{t} $}
   \ENDFOR 
   \STATE {\bfseries Output:} {\small $\boldsymbol{\theta}_{T}$  }
   \\ 
  %\hrulefill \\
  %  \vspace{1mm}\hrule  width0.45\textwidth \vspace{1mm}
 %\hspace{-5mm} *\hspace{0.1mm} {\small {Typical setups}: $\beta_1=0.9$, $\beta_2=0.999$, $\epsilon=1e-8$, $\eta_0=0.001$.  }   
\end{algorithmic}
\end{algorithm} 

\vspace{-2mm}
\section{Update procedure of Adam+}
\label{appendix:adamplus}
The parameter $\epsilon$ is added in the computation of $\boldsymbol{v}_t$ in (\ref{equ:Adam}) to verify if Adam+ has a small range of adaptive stepsizes than Adam. 
\label{appendix:adamplus}
\begin{algorithm}[h!]
   \caption{\small Adam+}
   \label{alg:Adamplus}
\begin{algorithmic}[1]
   \STATE {\small {\bfseries Input:} $\beta_1$, $\beta_2$,  $\eta_t$, $\epsilon > 0$ }
   \STATE {\small {\bfseries  Init.:} $\boldsymbol{\theta}_0\hspace{-0.5mm}\in\hspace{-0.5mm} \mathbb{R}^d$,  $\boldsymbol{m}_0 \hspace{-0.5mm}=\hspace{-0.5mm} 0$, $\boldsymbol{v}_{0}=0 \in \mathbb{R}^d$ }
   \FOR{\small $t=1, 2, \ldots, T$}
   \STATE \hspace{-0mm}{\small  $\boldsymbol{g}_t \leftarrow \nabla f({\boldsymbol{\theta}}_{t-1}) $  }
   \STATE \hspace{-0mm}{\small $\boldsymbol{m}_{t} \leftarrow \beta_1 \boldsymbol{m}_{t-1}  + (1-\beta_1) \boldsymbol{g}_t$ }
    \STATE \hspace{-0mm}{\small ${q}_{t}  \leftarrow \beta_2 q_{t-1} + (1-\beta_2) \boldsymbol{g}_{t}^2+\textcolor{blue}{\epsilon}$}
  % \STATE {\small $  {x}_t \leftarrow {x}_{t-1} \hspace{-0.6mm} - \hspace{-0.6mm}  \left\{\hspace{-1.5mm} \begin{array}{l}  \eta \frac{ 1}{(1-\beta_1^t)^q}  \frac{|{m}_t|^{q-1}\odot m_t}{(r_t/(1-\beta_2^t))^{q/p}+ \epsilon}    \\
  %\eta \frac{ (1-\beta_2^t))^{q/p} }{(1-\beta_1^t)^q}  \frac{|{m}_t|^{q-1}\odot m_t}{(r_t)^{q/p}+ \epsilon} 
    %\end{array} \right. $ }
   \STATE \hspace{-0mm}{\small $  \tilde{\boldsymbol{m}}_{t} \hspace{-0.6mm}\leftarrow \frac{\boldsymbol{m}_{t}}{1-\beta_1^{t}}\quad   \tilde{q}_{t} \leftarrow \frac{q_{t}}{1-\beta_2^{t}} $ }     
  \STATE {\small  $ \boldsymbol{\theta}_{t} \hspace{-0.6mm}\leftarrow \boldsymbol{\theta}_{t-1} -\frac{\eta_t }{\sqrt{\tilde{q}_{t}}+\epsilon} \tilde{\boldsymbol{m}}_{t} $}
   \ENDFOR 
   \STATE {\bfseries Output:} {\small $\boldsymbol{\theta}_{T}$  }
   \\ 
  %\hrulefill \\
  %  \vspace{1mm}\hrule  width0.45\textwidth \vspace{1mm}
 %\hspace{-5mm} *\hspace{0.1mm} {\small {Typical setups}: $\beta_1=0.9$, $\beta_2=0.999$, $\epsilon=1e-8$, $\eta_0=0.001$.  }   
\end{algorithmic}
\end{algorithm} 

\vspace{-2mm}
\section{Comparison of Adam+ and Adam for training VGG11 over CIAFR10 }
\label{appendix:Adamplus_exp}
The parameter setups for Adam+ and Adam can be found in Appendix~\ref{appendix:fig_setup}. It is clear from Fig.~\ref{fig:AdamPlus} that Adam+ does indeed have a smaller range of adaptive stepsizes than Adam. The above results are consistent with those in Fig.~\ref{fig:AdaBeliefMinus} for the comparison of AdaBelief and AdaBelief$^*$.

\begin{figure}[h!]
\centering
\includegraphics[width=60mm]{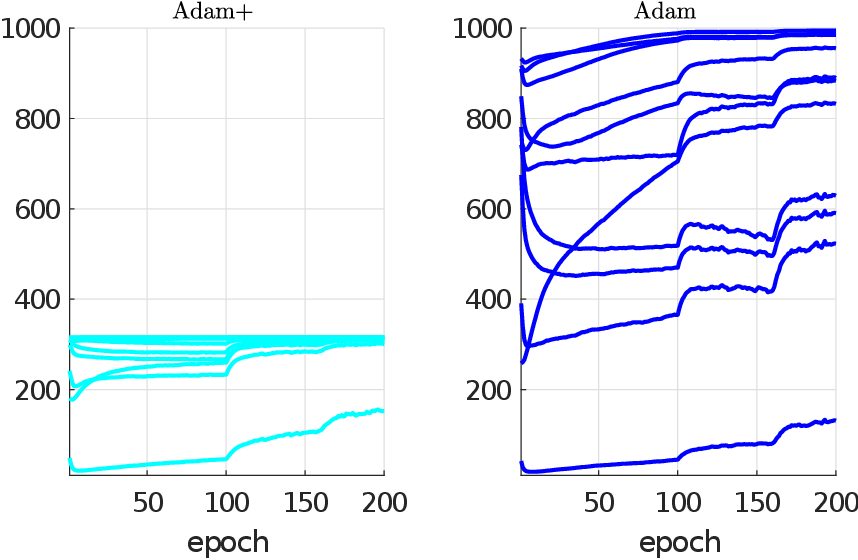}
\vspace*{-0.0cm}
\caption{\footnotesize{Comparison of layerwise average of adaptive stepsizes for the 11 neural layers by training VGG11 over CIFAR10 for 200 epochs. Adam+ is obtained by adding  $\epsilon$ in computation of $\boldsymbol{v}_t$ in (\ref{equ:Adam}).  The optimal $\epsilon$ is selected from a discrete set for Adam that gives the best validation accuracy.  } }
\label{fig:AdamPlus}
\vspace*{-0.3cm}
\end{figure}

\vspace{-2mm}
\section{Parameter Setups for Optimization Methods in  Fig.~\ref{fig:Aida_mean_compare}-\ref{fig:AdaBeliefMinus} and Fig.~\ref{fig:AdamPlus}}
\label{appendix:fig_setup}

The common stepsize $\eta_t$ is dropped by a factor 0.1 at 100 and 160 epochs.   The optimal parameter $\epsilon$ is searched over the set $\{10^{-2}, 10^{-3},\ldots, 10^{-8}\}$ for Adam and AdaBelief$^*$. 
\begin{table}[h!]
\label{tab:setup_fig}
\centering
\begin{tabular}{|c|c|c|}
\hline
\footnotesize{optimizer} & \footnotesize{fixed parameters} & \footnotesize{searched parameters}  \\ \hline
\hspace{-4.5mm} 
\footnotesize{Adam} \hspace{-4.5mm} & \hspace{-4.5mm} 
\footnotesize{$\begin{array}{l}(\eta_0,\beta_1, \beta_2) \\
=(0.001,0.9, 0.999)\end{array}$} \hspace{-5.5mm} & \hspace{-3.5mm}\footnotesize{$\begin{array}{c} \epsilon\in\{10^{-2}, 10^{-3}, \ldots,  10^{-8}\} \end{array}$} \hspace{-4.5mm} 
\\
\hline \hspace{-4.5mm} 
\footnotesize{AdaBelief$^*$} \hspace{-4.5mm} & \hspace{-4.5mm}
\footnotesize{$\begin{array}{l}(\eta_0,\beta_1, \beta_2) \\
=(0.001,0.9, 0.999)\end{array}$} \hspace{-5.5mm}  & \hspace{-3.5mm}  \footnotesize{$\begin{array}{c} \epsilon\in\{10^{-2}, 10^{-3}, \ldots,  10^{-8}\} \end{array}$} \hspace{-4.5mm} 
\\
\hline
\hspace{-4.5mm} 
\footnotesize{Adam+} \hspace{-4.5mm} & \hspace{-4.5mm}
\footnotesize{$\begin{array}{l}(\eta_0,\beta_1, \beta_2, \epsilon)\\
=(0.001, 0.9, 0.999, 10^{-8})\end{array}$} \hspace{-5.5mm}  & \hspace{-4.5mm} 
\\
\hline
\hspace{-4.5mm} 
\footnotesize{AdaBelief} \hspace{-4.5mm} & \hspace{-4.5mm} 
\footnotesize{$\begin{array}{l}(\eta_0,\beta_1, \beta_2, \epsilon)\\
=(0.001, 0.9, 0.999, 10^{-8})\end{array}$} \hspace{-4.5mm} & \hspace{-4.5mm}
\\
\hline
\hspace{-4.5mm} 
\footnotesize{$\begin{array}{c}\textrm{Aida} \\ (K=1)\end{array}$}  \hspace{-4.5mm} & \hspace{-4.5mm} \footnotesize{$\begin{array}{l}(\eta_0,\beta_1, \beta_2, \epsilon)\\
=(0.001, 0.9, 0.999, 10^{-8})\end{array}$} \hspace{-4.5mm} & \hspace{-4.5mm}
\\
\hline
\hspace{-4.5mm} 
\footnotesize{$\begin{array}{c}\textrm{Aida} \\ (K=2)\end{array}$} \hspace{-4.5mm}  &  \hspace{-4.5mm} \footnotesize{$\begin{array}{l}(\eta_0,\beta_1, \beta_2, \epsilon)\\
=(0.001, 0.9, 0.999, 10^{-8})\end{array}$} \hspace{-4.5mm} & \hspace{-4.5mm}
\\
\hline
\end{tabular}
\vspace{-1mm}
\end{table}

\vspace{-2mm}
\section{Parameter-setups for training different DNN models}
\label{appendix:parameterSetups}

\begin{table}[h!]
\caption{\small Parameter-setups  for training ResNet18 over ImageNet. The weight decay was set to $10^{-2}$.  \vspace{-0mm} } 
\label{tab:setup_imagenet}
\centering
\begin{tabular}{|c|c|c|}
\hline
\footnotesize{optimizer} & \footnotesize{fixed parameters} & \footnotesize{searched parameters}  \\ \hline
\hspace{-4.5mm} 
\footnotesize{AdaBelief} \hspace{-4.5mm} & \hspace{-4.5mm} 
\footnotesize{$\begin{array}{l}(\eta_0,\beta_1, \beta_2)\\
=(0.001, 0.9, 0.999)\end{array}$} \hspace{-4.5mm} & \hspace{-2.5mm} 
\footnotesize{$\epsilon=\{10^{-8}, 10^{-9}, 10^{-10}\}$}
\\
\hline
\hspace{-4.5mm} 
\footnotesize{$\begin{array}{c}\textrm{Aida} \\ (K=2)\end{array}$} \hspace{-4.5mm}  &  \hspace{-4.5mm} \footnotesize{$\begin{array}{l}(\eta_0,\beta_1, \beta_2, \epsilon)\\
=(0.001, 0.9, 0.999, 10^{-9})\end{array}$} \hspace{-4.5mm} & \hspace{-4.5mm}
\\
\hline
\end{tabular}
\vspace{-1mm}
\end{table}

%As stated at the beginning of Section-\textbf{Experiments}, the open source ``https://github.com/jadore801120/attention-is-all-you-need-pytorch" was adopted for training a Transformer. All the remaining tasks rely on the second open source 
%``https://github.com/juntang-zhuang/Adabelief-Optimizer", which is in fact for evaluating AdaBelief of \cite{Zhuang20Adabelief}. The four tables below specify the selection of parameters for each optimizer in four DNN training tasks.  

%\textcolor{blue}{In addition, we also evaluated the AdaBound method from \cite{Luo19AdaBound}. The tables below also include the parameter setups for AdaBound. The experimental results for AdaBound are presented in Appendix~G. }  

\begin{table*}[h!]
\caption{\footnotesize Parameter-setups  for training a Transformer. The weight decay for AdamW was set to $5e-4$ while the weight decay for all other algorithms was set to 0.0.  \vspace{-2mm} } 
\label{tab:setup_transformer}
\centering
\begin{tabular}{|c|c|c|}
\hline
\footnotesize{optimizer} & \footnotesize{fixed parameters} & \footnotesize{searched parameters}  \\ \hline
\footnotesize{AdaBound} & \footnotesize{$(\eta_0,\beta_1,\beta_2,\gamma)=(0.001,0.9,0.98,0.001)$} & \footnotesize{$\begin{array}{c}\epsilon\in\{1e-6, 1e-7, \ldots,  1e-12\} \\ 
\textrm{final}\_\textrm{stepsize}\in \{0.1, 0.01,0.001\}\end{array}$}
\\
\hline
\footnotesize{Yogi} & \footnotesize{$(\eta_0,\beta_1,\beta_2)=(0.001,0.9,0.98)$} & \footnotesize{$\epsilon\in\{1e-2, 1e-3, \ldots,  1e-8\}$}
\\
\hline
\footnotesize{SGD} & \footnotesize{momentum=0.9} &
\footnotesize{$\eta_0\in\{1.0, 0.1, 0.01, 0.001\}$ }
\\
\hline
\footnotesize{RAdam} & 
\footnotesize{$(\eta_0,\beta_1, \beta_2)=(0.001, 0.9, 0.98)$} & \footnotesize{$\epsilon\in\{1e-6, 1e-7, \ldots,  1e-12\}$}
\\
\hline
\footnotesize{MSVAG} & \footnotesize{$(\eta_0,\beta_1,\beta_2)=(0.001,0.9,0.98)$} & \footnotesize{$\epsilon\in\{1e-6, 1e-7, \ldots,  1e-12\}$}
\\
\hline
\footnotesize{Fromage} & & \footnotesize{$\eta_0\in\{0.1, 0.01, 0.001, 0.0001\}$}
\\
\hline
\footnotesize{Adam} & 
\footnotesize{$(\eta_0,\beta_1, \beta_2)=(0.001, 0.9, 0.98)$} & \footnotesize{$\epsilon\in\{1e-6, 1e-7, \ldots,  1e-12\}$}
\\
\hline
\footnotesize{AdamW} & 
\footnotesize{$(\eta_0,\beta_1, \beta_2)=(0.001, 0.9, 0.98)$} & \footnotesize{$\epsilon\in\{1e-6, 1e-7, \ldots,  1e-12\}$}
\\
\hline
\footnotesize{AdaBelief} & 
\footnotesize{$(\eta_0,\beta_1, \beta_2)=(0.001, 0.9, 0.98)$} & \footnotesize{$\epsilon\in\{1e-8, 1e-10, \ldots,  1e-16\}$}
\\
\hline
\hline
\footnotesize{Aida(K=1)(\textbf{our})} & 
\footnotesize{$(\eta_0,\beta_1, \beta_2, \epsilon)=(0.001, 0.9, 0.98, 1e-16)$} & 
\\
\hline
\footnotesize{Aida(K=2)(\textbf{our})} & \footnotesize{$(\eta_0,\beta_1, \beta_2, \epsilon)=(0.001, 0.9, 0.98, 1e-16)$} & 
\\
\hline
\end{tabular}
\vspace{-1mm}
\end{table*}\

\begin{table*}[h!]
\caption{\footnotesize Parameter-setups for training LSTMs. The weight decay for every algorithm was set to $1.2e-6$.   \vspace{-2mm} } 
\label{tab:setup_LSTM}
\centering
\begin{tabular}{|c|c|c|}
\hline
\footnotesize{optimizer} & \footnotesize{fixed parameters} & \footnotesize{searched parameters}  \\ \hline
\\ \hline
\footnotesize{AdaBound} & \footnotesize{$(\beta_1, \beta_2,\gamma)=(0.9, 0.999,0.001)$} & \footnotesize{$\begin{array}{c} 
\eta_0 \in \{0.01, 0.001\} \\
\epsilon\in\{1e-6, 1e-8, \ldots, 1e-12\} \\
\textrm{final}\_\textrm{stepsize}\in \{0.1, 3, 30\}\end{array}$} \\
\hline
\footnotesize{Yogi} & \footnotesize{$(\beta_1, \beta_2)=(0.9, 0.999)$} & \footnotesize{$\begin{array}{c}\eta_0\in \{0.01, 0.001\} \\  \epsilon\in\{1e-2, 1e-3, 1e-4, 1e-5\} \end{array}$}
\\
\hline
\footnotesize{SGD} & \footnotesize{momentum=0.9} &
\footnotesize{$\eta_0\in\{30, 3, 1, 0.1\}$ }
\\
\hline
\footnotesize{RAdam} & 
\footnotesize{$(\beta_1, \beta_2)=(0.9, 0.999)$} & \footnotesize{$\begin{array}{c}\eta_0\in \{0.01, 0.001\} \\  \epsilon\in\{1e-6, 1e-8, 1e-10,  1e-12\} \end{array}$}
\\
\hline
\footnotesize{MSVAG} & \footnotesize{$(\beta_1, \beta_2)=(0.9, 0.999)$} & \footnotesize{$\begin{array}{c}\eta_0\in \{30, 1, 0.01, 0.001\} \\  \epsilon\in\{1e-6, 1e-8, 1e-10, 1e-12\} \end{array}$}
\\
\hline
\footnotesize{Fromage} & & \footnotesize{$\eta_0\in\{0.1, 0.01, 0.001\}$}
\\
\hline
\footnotesize{Adam} & 
\footnotesize{$(\beta_1, \beta_2)=(0.9, 0.999)$} & \footnotesize{$\begin{array}{c}\eta_0\in \{0.01, 0.001\} \\  \epsilon\in\{1e-6, 1e-8, 1e-10,  1e-12\} \end{array}$}
\\
\hline
\footnotesize{AdamW} & 
\footnotesize{$(\beta_1, \beta_2)=(0.9, 0.999)$} & \footnotesize{$\begin{array}{c}\eta_0\in \{0.01, 0.001\} \\  \epsilon\in\{1e-6, 1e-8, 1e-10,  1e-12\} \end{array}$}
\\
\hline
\footnotesize{AdaBelief} & 
\footnotesize{$(\beta_1, \beta_2)=(0.9, 0.999)$} & \footnotesize{$\begin{array}{c}\eta_0\in \{0.01, 0.001\} \\  \epsilon\in\{1e-8, 1e-10, \ldots,  1e-16\} \end{array}$}
\\
\hline
\footnotesize{Aida(K=1)} (\textbf{our}) & 
\footnotesize{$(\eta_0,\beta_1, \beta_2, \epsilon)=(0.001, 0.9, 0.999, 1e-16)$} & 
\\
\hline
\footnotesize{Aida(K=2)} (\textbf{our}) & \footnotesize{$(\eta_0,\beta_1, \beta_2, \epsilon)=(0.001, 0.9, 0.999, 1e-16)$} & 
\\
\hline
\end{tabular}
\vspace{-1mm}
\end{table*}

\begin{table*}[h!]
\caption{\footnotesize Parameter-setups  for training WGAN-GP.  The weight decay for AdamW was set to $5e-4$ while the weight decay for all other algorithms was set to 0.0.  \vspace{-2mm} } 
\label{tab:setup_WGANGP}
\centering
\begin{tabular}{|c|c|c|}
\hline
\footnotesize{optimizer} & \footnotesize{fixed parameters} & \footnotesize{searched parameters}  \\ \hline
\footnotesize{AdaBound} & \footnotesize{$(\eta_0,\beta_1, \beta_2,\gamma)=(0.0002,0.5, 0.999,0.001)$} & \footnotesize{$\begin{array}{c}  \epsilon\in\{1e-2, 1e-4, \ldots, 1e-10\} \\
\textrm{final}\_\textrm{stepsize}\in \{0.1, 0.01\}\end{array}$}
\\
\hline
\footnotesize{Yogi} & \footnotesize{$(\eta_0,\beta_1, \beta_2)=(0.0002,0.5, 0.999)$} & \footnotesize{$\begin{array}{c}  \epsilon\in\{1e-2, 1e-3, 1e-4, 1e-5\} \end{array}$}
\\
\hline
\footnotesize{SGD} &  &
\footnotesize{$\begin{array}{c}
\textrm{momentum} = \{0.3, 0.5, 0.9\}
\\\eta_0\in\{0.1, 0.02, 0.002, 0.0002\}\end{array}$ }
\\
\hline
\footnotesize{RAdam} & 
\footnotesize{$(\eta_0,\beta_1, \beta_2)=(0.0002,0.5, 0.999)$} & \footnotesize{$\begin{array}{c} \epsilon\in\{1e-4, 1e-6,\ldots,  1e-14\} \end{array}$}
\\
\hline
\footnotesize{MSVAG} & \footnotesize{$(\beta_1, \beta_2)=(0.5, 0.999)$} & \footnotesize{$\begin{array}{c}\eta_0\in \{0.1, 0.02, 0.002, 0.0002\} \\  \epsilon\in\{1e-4, 1e-6, \ldots, 1e-14\} \end{array}$}
\\
\hline
\footnotesize{Fromage} & & \footnotesize{$\eta_0\in\{0.1, 0.01, 0.001\}$}
\\
\hline
\footnotesize{Adam} & 
\footnotesize{$(\eta_0,\beta_1, \beta_2)=(0.0002,0.5, 0.999)$} & \footnotesize{$\begin{array}{c} \epsilon\in\{1e-4, 1e-6, \ldots,  1e-14\} \end{array}$}
\\
\hline
\footnotesize{AdamW} & 
\footnotesize{$(\eta_0, \beta_1, \beta_2)=(0.0002,0.5, 0.999)$} & \footnotesize{$\begin{array}{c} \epsilon\in\{1e-4, 1e-6, \ldots,  1e-14\} \end{array}$}
\\
\hline
\footnotesize{AdaBelief} & 
\footnotesize{($\eta_0,\beta_1, \beta_2, \epsilon)=(0.0002, 0.5, 0.999, 1e-12)$} & 
\\
\hline
\footnotesize{Aida(K=1)} (\textbf{our}) & 
\footnotesize{($\eta_0,\beta_1, \beta_2, \epsilon)=(0.0002, 0.5, 0.999, 1e-12)$} & 
\\
\hline
\footnotesize{Aida(K=2)} (\textbf{our}) & \footnotesize{($\eta_0,\beta_1, \beta_2, \epsilon)=(0.0002, 0.5, 0.999, 1e-12)$} & 
\\
\hline
\end{tabular}
\vspace{-1mm}
\end{table*}

\begin{table*}[h!]
\caption{\footnotesize Parameter-setups  for training VGG and ResNet models over CIFAR10 and CIFAR100.  The weight decay for AdamW was set to $0.01$ while the weight decay for all other algorithms was set to $5e-4$.   \vspace{-2mm} } 
\label{tab:setup_VGGResNet}
\centering
\begin{tabular}{|c|c|c|}
\hline
\footnotesize{optimizer} & \footnotesize{fixed-parameters}  & \footnotesize{searched-parameters} \\ \hline
\footnotesize{AdaBound} & \footnotesize{$(\eta_0,\beta_1, \beta_2,\gamma)=(0.001,0.9, 0.999, 0.001)$} & \footnotesize{
$\begin{array}{c}\epsilon\in \{1e-2,1e-3, \ldots, 1e-8\}
\\ \textrm{final}\_\textrm{stepsize} \in\{0.1, 0.01\} \end{array}$ }
\\
\hline
\footnotesize{Yogi} & \footnotesize{$(\eta_0,\beta_1, \beta_2,)=(0.001,0.9, 0.999$)} & \footnotesize{$\epsilon\in \{1e-1,1e-2, 1e-3\}$ }
\\
\hline
\footnotesize{SGD} &  
\footnotesize{$(\textrm{momentum}, \eta_0)=(0.9,0.1)$}
\\
\hline
\footnotesize{RAdam} & 
\footnotesize{$(\eta_0,\beta_1, \beta_2)=(0.001,0.9, 0.999)$} &  \footnotesize{$\epsilon\in \{1e-2, 1e-3, \ldots, 1e-8\}$}
\\
\hline
\footnotesize{MSVAG} & \footnotesize{$(\eta_0,\beta_1, \beta_2)=(0.1,0.9, 0.999)$} & \footnotesize{$\epsilon\in \{1e-2, 1e-2,\ldots, 1e-8\}$}
\\
\hline
\footnotesize{Fromage} & &
\footnotesize{$\eta_0\in \{0.1, 0.01, 0.001\}$}
\\
\hline
\footnotesize{Adam} & 
\footnotesize{$(\eta_0,\beta_1, \beta_2)=(0.001,0.9, 0.999)$} & \footnotesize{$\epsilon\in\{1e-2,1e-3,\ldots, 1e-8\}$}
\\
\hline
\footnotesize{AdamW} & 
\footnotesize{$(\eta_0,\beta_1, \beta_2)=(0.001,0.9, 0.999)$} & \footnotesize{$\epsilon\in\{1e-2,1e-3,\ldots, 1e-8\}$}
\\
\hline
\footnotesize{AdaBelief} & 
\footnotesize{$(\eta_0,\beta_1, \beta_2)=(0.001,0.9, 0.999)$} & \footnotesize{$\epsilon\in\{1e-8, 1e-9\}$}
\\
\hline
\footnotesize{Aida(K=1) ($\textbf{our}$) }  & 
\footnotesize{$(\eta_0,\beta_1, \beta_2, \epsilon)=(0.001,0.9, 0.999, 1e-8)$} &
\\
\hline
\footnotesize{Aida(K=2) ($\textbf{our}$)}  & \footnotesize{$(\eta_0,\beta_1, \beta_2, \epsilon)=(0.001,0.9, 0.999, 1e-9)$}  &
\\
\hline
\end{tabular}
\vspace{-1mm}
\end{table*}

%\newpage

\end{document}